\documentclass[journal]{IEEEtran}

\usepackage[pdftex]{graphicx}
\usepackage{cite}
\usepackage{amsopn}
\usepackage{booktabs} 
\usepackage{url}
\usepackage{color}
\usepackage{multirow}
\usepackage{xcolor}
\usepackage[switch]{lineno}
\usepackage{graphicx}
\usepackage{subfigure}
\usepackage{url}
\usepackage{epstopdf}
\usepackage{soul}
\usepackage{xcolor}
\usepackage{amsmath}
\usepackage{textcomp}
\usepackage{verbatim}
\usepackage{bm}
\usepackage{hyperref}
\usepackage{amssymb}

\begin{document}

\title{Aerial Images Meet Crowdsourced Trajectories: A New Approach to Robust Road Extraction}

\author{Lingbo Liu,
        Zewei Yang,
        Guanbin Li,
        Kuo Wang,
        Tianshui Chen
        and Liang Lin, {\textit{Senior Member, IEEE}}
\thanks{This work was supported in part by the Guangdong Basic and Applied Basic Research Foundation under Grant No.2020B1515020048, in part by the National Natural Science Foundation of China under Grant No.61976250, No.U1811463 and No.61836012, and in part by the Guangzhou Science and Technology Project under Grant No.202102020633. (\textit{Corresponding Author: Liang Lin.})}
\thanks{L. Liu, G. Li, K. Wang and L. Lin are with the School of Computer Science and Engineering, Sun Yat-Sen University, China, 510000 (e-mail: liulingb@mail2.sysu.edu.cn; liguanbin@mail.sysu.edu.cn; wangk229@mail2.sysu.edu.cn; linliang@ieee.org).}
\thanks{Z. Yang is with the School of Mathematics, Sun Yat-Sen University, China, 510000 (e-mail: yangzw7@mail2.sysu.edu.cn).}
\thanks{T. Chen is with The Guangdong University of Technology, Guangzhou, China, 510000 (e-mail: tianshuichen@gmail.com).}
}

\markboth{IEEE Transactions on Neural Networks and Learning Systems}
{Liu \MakeLowercase{\textit{et al.}}: Aerial Images Meet Crowdsourced Trajectories}

\maketitle

\begin{abstract}
  Land remote sensing analysis is a crucial research in earth science. In this work, we focus on a challenging task of land analysis, i.e., automatic extraction of traffic roads from remote sensing data, which has widespread applications in urban development and expansion estimation. Nevertheless, conventional methods either only utilized the limited information of aerial images, or simply fused multimodal information (e.g., vehicle trajectories), thus cannot well recognize unconstrained roads. To facilitate this problem, we introduce a novel neural network framework termed Cross-Modal Message Propagation Network (CMMPNet), which fully benefits the complementary different modal data (i.e., aerial images and crowdsourced trajectories). Specifically, CMMPNet is composed of two deep Auto-Encoders for modality-specific representation learning and a tailor-designed Dual Enhancement Module for cross-modal representation refinement. In particular, the complementary information of each modality is comprehensively extracted and dynamically propagated to enhance the representation of another modality.
  Extensive experiments on three real-world benchmarks demonstrate the effectiveness of our CMMPNet for robust road extraction benefiting from blending different modal data, either using image and trajectory data or image and Lidar data. From the experimental results, we observe that the proposed approach outperforms current state-of-the-art methods by large margins. Our source code is resealed on the project page {\color{blue}\url{http://lingboliu.com/multimodal_road_extraction.html}}.
\end{abstract}


\IEEEpeerreviewmaketitle
\section{Introduction}\label{sec:introduction}

\IEEEPARstart{E}{arth} science \cite{von1988theory,national2001basic} is a complex and huge subject that has been researched for decades or even centuries. As a subbranch of geoscience, geoinformatics \cite{keller2011geoinformatics} recently has received increasing interests with the rapid development of satellite and computer technologies. Accurately obtaining land surface information (e.g., trees, lakes, buildings, roads, and so on) from remote sensing data can help us to better understand our earth. Among these objects, traffic roads are very difficult to recognize, since they are threadlike and unimpressive in aerial images. To promote land analysis, in this work we aim to recognize traffic roads automatically from remote sensing data. Such a geoinformatics task not only facilitates a series of practical applications \cite{zhang2015efficient,wang2020attention,wang2020multitask} for urban development, but also helps to estimate the urban expansion trend to analyze potential impacts of human activities on earth lands.

\begin{figure}
\centerline{
\includegraphics[width=0.925\columnwidth]{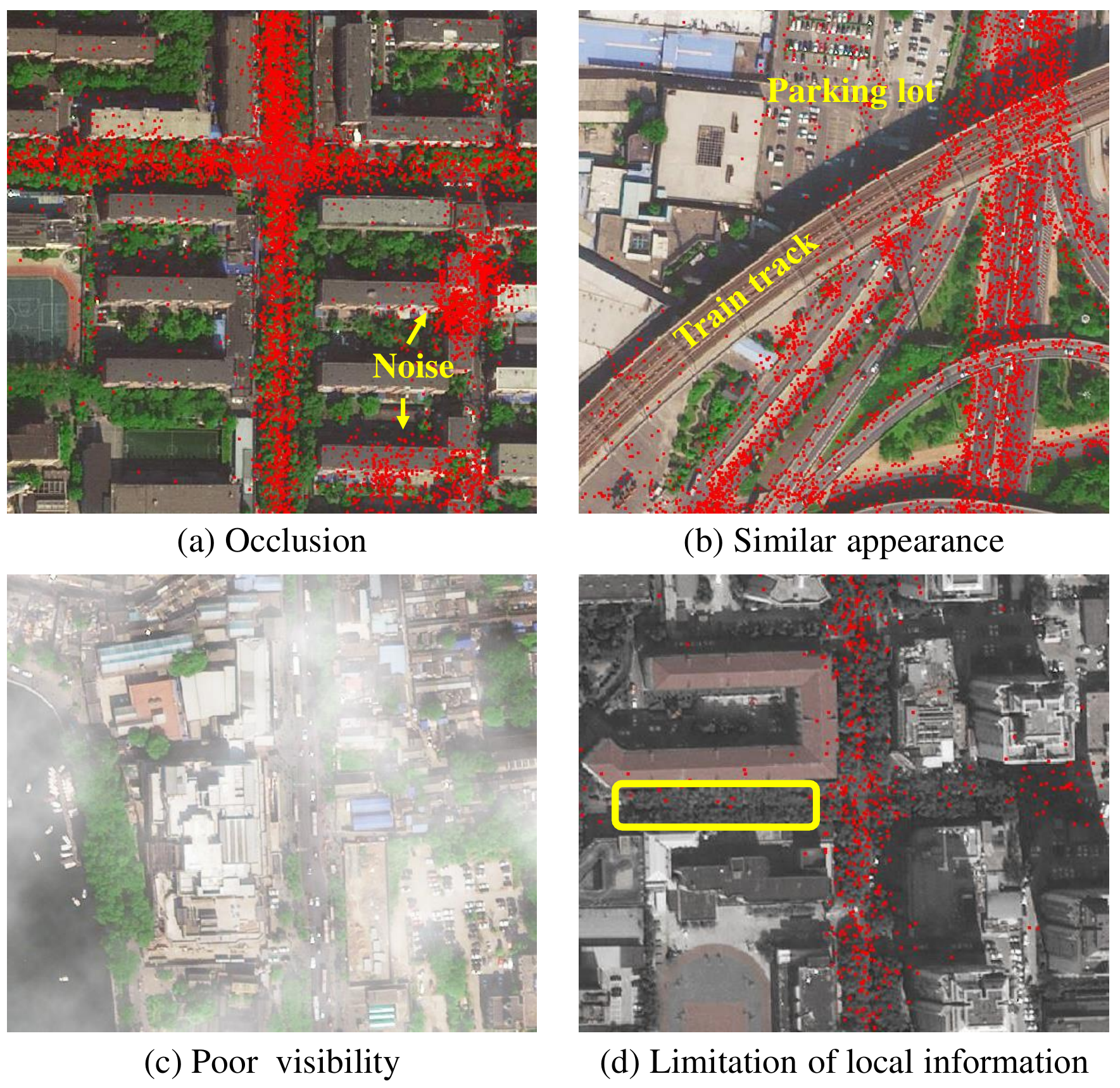}
}
\vspace{-3mm}
   \caption{{\bf{(a)}} Traffic roads are usually occluded by trees. Although crowdsourced trajectories can help discover roads, excessive noises are also introduced. {\bf{(b)}} Train tracks and traffic roads have similar appearances, thus it is hard to distinguish them only using visual cues. When only using trajectories, some parking lots are easily mistaken for roads. {\bf{(c)}} It's difficult to directly recognize roads from aerial images when the studied city has poor visibility in fog/haze weather. {\bf{(d)}} Only using local information, we may fail to recognize some road regions that are heavily occluded and have very few trajectories, as shown in the yellow box.}
\vspace{0mm}
\label{fig:challenge}
\end{figure}

In literature, numerous algorithms have been proposed to extract traffic roads from aerial images. Most early works \cite{wang2005extracting,movaghati2010road,shi2013integrated} extracted handcrafted features (e.g., texture and contour) and applied shallow models (e.g., Support Vector Machine \cite{suykens1999least} and Markov Random Field \cite{li2009markov}) to recognize road regions. Recently, deep convolutional networks have become the mainstream in this field and achieved remarkable progresses \cite{cheng2017automatic,buslaev2018fully,lu2019multi} due to their great capacities of representation learning.
However, aerial image-based traffic road extraction remains a very challenging problem, especially in the face of the following circumstances. {\bf{First}}, some roads are extremely occluded by trees, as shown in Fig. \ref{fig:challenge}-(a). Relying solely on visual information, these roads are hard to be detected from aerial images. {\bf{Second}}, some infrastructures (e.g., train tracks, building tops, and river walls) have similar appearances of traffic roads, as shown in Fig. \ref{fig:challenge}-(b). Without extra information, it is hard to distinguish roads from these structures, which may result in false negatives and false positives. {\bf{Third}}, in some bad meteorological conditions (e.g., thick fog/haze), it's very difficult to recognize traffic roads due to poor visibility, as shown in Fig. \ref{fig:challenge}-(c).
Nevertheless, road maps have low tolerance for errors, since incorrect routes would seriously affect the transportation's operation efficiency. Therefore, some robust methods are desired to accurately extract traffic roads.

Fortunately, we observe that some data for non-visual modalities, such as vehicle trajectories, can also help discover traffic roads. Intuitively, a region with a large number of trajectories is likely to be a road segment \cite{rogers1999mining,schroedl2004mining,davies2006scalable}. In recent years, vehicle ownership has grown dramatically and most vehicles have been equipped with GPS devices, which greatly increases the availability of large-scale trajectory datasets and boosts the feasibility of trajectory-based road extraction.
Despite substantial progress \cite{liu2012mining,biagioni2012map}, this research direction still suffers from many challenges. {\bf{First}}, crowdsourced trajectories have excessive noises (e.g., positioning drift) caused by the uneven quality of GPS devices, as shown in Fig. \ref{fig:challenge}-(a). Although various preprocessing techniques (e.g., clustering and K-nearest neighbors) were used \cite{wang2013crowdatlas,shan2015cobweb,karagiorgou2017layered}, the noise problem has not been well solved. {\bf{Second}}, some non-road areas, such as the parking lot in Fig. \ref{fig:challenge}-(b), also have lots of trajectories and they are easily mistaken for roads without auxiliary information. Most conventional works \cite{shi2009automatic,chen2016city,yang2018method,ruan2020learning} have not explicitly distinguished these areas. {\bf{Third}}, previous trajectory-based methods mainly extracted the topologies of road networks. Because of mass erratic trajectories, it is difficult to obtain the accurate width of roads, which can be easily computed in high-resolution aerial images.

In general, image-based methods and trajectory-based methods have individual strengths and weaknesses. It is very natural to incorporate aerial images and crowdsourced trajectories to extract traffic roads robustly. However, there are very limited works \cite{sun2019leveraging,li2019fusing} that simultaneously utilized the two modalities mentioned above. Moreover, these works directly fed the concatenation of aerial images and rendered maps of trajectories or their features into convolutional neural networks, which is a suboptimal strategy for multimodal fusion. Recently, Wu \textit{et al.} \cite{wu2020deepdualmapper} designed a gated fusion module to fuse multimodal features, but not refine features mutually, thus the complementarities of images and trajectories have not been fully exploited.
Furthermore, all above-mentioned methods performed road extraction only with local features/information, thus may fail to recognize some road regions that are heavily occluded and meanwhile have very few trajectories, such as the yellow box in Fig. \ref{fig:challenge}-(d). When considering all information of the whole image and trajectories, we can correctly infer that this region is a road segment. Therefore, both the local and global information should be explored for traffic road extraction.

To facilitate road extraction, we propose a novel framework termed Cross-Modal Message Propagation Network (CMMPNet), which fully explores the complementarities between aerial images and vehicle crowdsourced trajectories. Specifically, our CMMPNet is composed of: {\bf\color{red}{(i)}} two deep AutoEncoders for modality-specific feature learning, in which one takes an aerial image as input and the other one uses the rendered trajectory heat-map, and {\bf\color{red}{(ii)}} a Dual Enhancement Module (DEM) that refines the features of different modalities mutually with a message passing mechanism. In particular, our DEM propagates both the local detail information and global structural information dynamically with two progress propagators.
{\bf{First}}, a Non-Local Message Propagator extracts the local and global messages embedded in the features of each modality, which are utilized to refine the features of another modality. Thereby, image features and trajectories features can be enhanced mutually. Moreover, the limitation of local information is also well eliminated. {\bf{Second}}, a Gated Message Propagator employs gate functions to dynamically determine the final propagated messages, so that the beneficial messages are transmitted and the interferential messages (e.g., visual cues of train tracks and the noises of trajectories) are abandoned.
For further improving the robustness, our DEM is integrated into different layers of CMMPNet to enhance the image features and trajectory features hierarchically. Finally, the last outputs of two AutoEncoders are concatenated to accurately predict the high-resolution traffic road maps.

The proposed CMMPNet has three appealing properties.
{\bf{First}}, through refining modality-specific features mutually, our method can better explore the complementarities of aerial images and crowdsourced trajectories, compared with previous works that directly taken their concatenation as input or simply fused their features.
{\bf{Second}}, thanks to the tailor-designed DEM, our method is more robust to extract traffic roads. With the aid of visual information, some useless and noisy trajectories can be effectively eliminated, while occluded roads are easily discovered with the trajectory information and some delusive non-road regions are also well distinguished.
{\bf{Third}}, it is worth noting that our method is very general for robust road extraction by utilizing multimodal information. Furthermore, CMMPNet can also be generalized to combine image and lidar data for road extraction. Extensive comparisons on three real-world benchmarks two for image and trajectory data and the other for image and lidar data) demonstrate the advantage of our proposed method.
In summary, this paper makes the following contributions:

\begin{itemize}
\item It proposes a novel Cross-Modal Message Propagation Network for land remote sensing analysis, which extracts traffic roads robustly by explicitly capturing the complementarities among different modal data.
\item It introduces a Dual Refinement Module for multimodal representation learning, where the complementary information of each modality is dynamically propagated to effectively enhance other modal features based on the message passing mechanism.
\item It presents sufficient experiments and comparisons on three multimodal benchmarks for showing the superiority and generalization of our approach against existing state-of-the-art methods.
\end{itemize}

The rest of this paper is organized as follows. First, we review some related works of earth science research and traffic road extraction in Section \ref{sec:review}. We then provide some preliminaries in Section \ref{sec:preliminary} and introduce the proposed CMMPNet in Section \ref{sec:CMMP}. Extensive evaluations and generalization analysis are conducted in Section \ref{sec:experiment} and in Section \ref{sec:lidar_experiment}. Finally, we conclude this paper and discuss future works in Section \ref{sec:conclusion}.

\section{Related Works}\label{sec:review}
\subsection{Earth Science Research}
Earth science \cite{von1988theory,national2001basic} is a crucial subject that studies the physical, chemical, and biological characterizations of our earth for better understanding various physical phenomena and natural systems. Earth science is also a complex subject and it contains a lot of research branches \cite{Earthscience}. For instance, meteorologists \cite{fine2009authors} study the atmosphere for dangerous storm warnings and hydrologists \cite{birylo2015creation} examine hydrosphere for flood warnings. Seismologists \cite{kagan2000probabilistic} study earthquakes and forecast where they will strike, while geologists \cite{clark1999spectroscopy} study rocks and help to locate useful minerals. Among all the subbranches of geoscience, geoinformatics \cite{keller2011geoinformatics} recently has attracted widespread interests with the rapid development of satellite and computer technologies, since it can greatly facilitate other research branches, e.g., monitoring storm/flood from remote sensing data and forecasting their evolutionary trend. In this work, we inherit the research content of geoinformatics and apply computer technologies to land remote sensing analysis, e.g, extracting the traffic road network from aerial images and some complementary modalities. This problem has important applications in transportation navigation and public management. Moreover, we can also compare the road networks at different times and estimate the urban expansion tendency, thereby analyzing the potential impacts of human activities on earth lands.

\subsection{Traffic Road Extraction}

As a crucial foundation in intelligent transportation systems, automatic road extraction has been studied for decades \cite{wang2016review}. On the basis of the modality of input data, previous approaches can be divided into four categories and we would investigate the related works of each category.

\subsubsection{Aerial Image-based Road Extraction}
In industrial communities, a large number of high-quality aerial images can be accessed easily, with the rapid development of remote sensing imaging technologies equipped in artificial satellites \cite{zhang2020local,lin2020crpn}. Numerous methods were proposed to extract traffic roads from these aerial images. Early works \cite{hinz2003automatic,anil2010novel,chaudhuri2012semi,leninisha2015water} usually fed hand-crafted features (e.g., texture and contour) into shallow models (e.g., deformable model and Markov Random Field) to recognize road regions. However, most of them only worked in constraint scenarios.
In recent years, due to the great capacity for representation learning, deep neural networks \cite{lecun2015deep} have become the mainstream in this field. For instance, Cheng \textit{et al.} \cite{cheng2017automatic} proposed a cascaded end-to-end convolutional neural network to cope with the road detection and centerline extraction simultaneously with two cascaded CNN. Zhang \textit{et al.} \cite{zhang2018road} developed a semantic segmentation neural network, which combined the residual learning and U-Net to extract road areas. Zhou \textit{et al.} \cite{zhou2018d} utilized dilation convolutions to enlarge the receptive field of Linknet \cite{chaurasia2017linknet} and then employed this enhanced model to extract road regions from high-resolution aerial images. Fu \textit{et al.} \cite{fu2017classification} predicted the category of each pixel with a multi-scale Fully Convolutional Network and refined the output density map with a Conditional Random Fields post-processing.
Despite substantial progress, they may still fail in complex scenarios, especially in the face of extreme occlusions. As analyzed above, it's very difficult to perfectly extract traffic roads only with the visual information of aerial images. Therefore, more complementary information should be delved from other modalities for facilitating road extraction.

\begin{figure*}[t]
  \begin{center}
     \includegraphics[width=1.5\columnwidth]{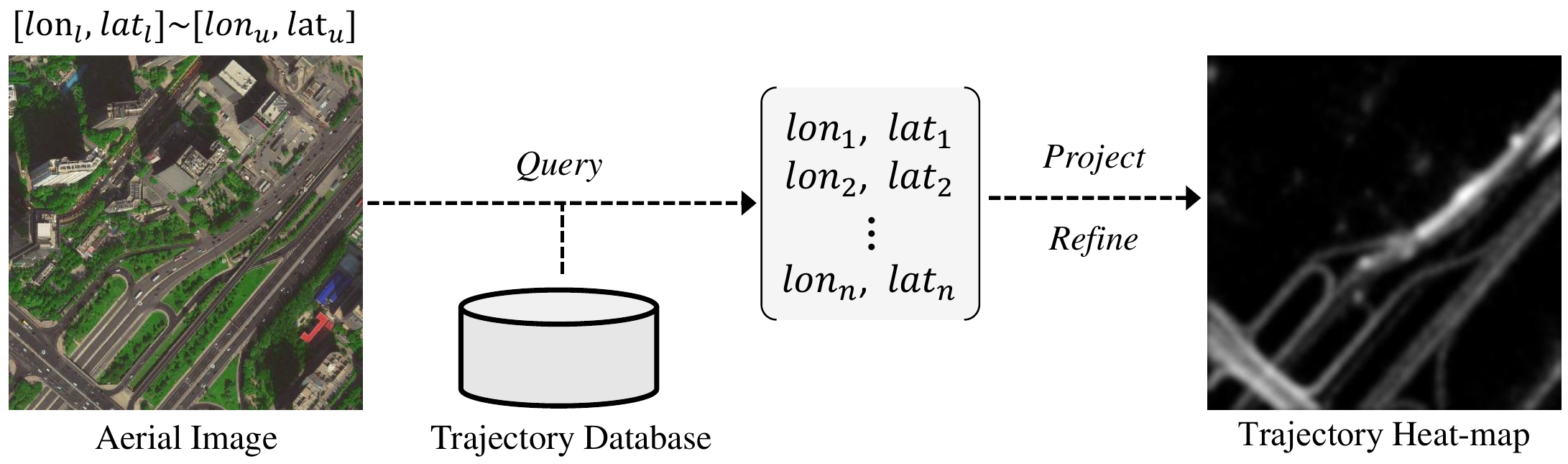}
  \vspace{-5mm}
  \end{center}
   \caption{Illustration of trajectory heat-map generation. Given an aerial image, we first query all trajectory samples in the corresponding geographical region and then generate a 2D trajectory heat-map by counting the number of samples projected at every pixel. Finally, a logarithm-based normalized function and a Gaussian kernel filter are applied to refine the trajectory heat-map.}
\vspace{0mm}
\label{fig:render_map}
\end{figure*}

\subsubsection{Trajectory-based Road Extraction}
Intuitively, a geographical region with mass vehicle trajectories is likely to be a road area. Based on this observation, some researchers have attempted to recognize traffic roads from crowdsourced trajectories. Since trajectory data has excessive noise, most previous works focused on how to eliminate the GPS noises and uncertainties. Conventional methods can be divided into three categories.
The first category is clustering-based models \cite{edelkamp2003route,chen2016city,stanojevic2018robust}. In these works, the task of road extraction is formulated as a network alignment optimization problem where both the nodes and edges of road networks have to be inferred. Specifically, nodes or short edges are first identified from raw GPS points with spatial clustering algorithms and then connected to form the final road networks.
The second category is trace-merging based models \cite{cao2009gps,niehoefer2009gps}, which either merge each trajectory to an existing road segment or generate a new segment if no existing segment is matching. The third category is KDE-based methods \cite{biagioni2012map,wang2015efficient}, which first apply Kernel Density Estimation \cite{terrell1992variable} to convert trajectories into a density map for reducing the influences of noise, and then employs image processing techniques to extract road centerlines. Recently, deep neural networks have also been applied to this task. For instance, Cheng \textit{et al.} \cite{ruan2020learning} proposed a deep learning-based map generation framework, which extracts features from trajectories in both spatial view and transition view to infer road centerlines.
Although various techniques are used, GPS noises still can't be well eliminated and the extracted road networks are far from satisfactory, due to the limited information of crowdsourced trajectories.

\subsubsection{Lidar-based Road Extraction}
Compared with aerial images, Light Detection And Ranging (Lidar) data have two specialties. First, Lidar data contains depth or distance information. Second, different objects (e.g., buildings, trees, and roads) have different reflectivity to the laser. Because of these specialties, roads are mostly defined by flatness in the aerial viewpoint, which can help to distinguish the road proposals from buildings and trees.
In literature, there also exist some algorithms \cite{clode2004automatic,hui2016road,zhang2010lidar} that identified traffic road from Lidar data. For instance, Hu et al. \cite{hu2014road} first filtered the non-ground LiDAR points and then detected road centerlines from the remaining ground points. After obtaining the ground intensity images, Zhao and You \cite{hu2014road} designed structure templates to search for roads and determined road widths and orientations with a subsequent voting scheme. Despite some progress, Lidar-based road extraction remains a very challenging problem and existing methods perform poorly in complex scenarios, suffering from the sparsity of Lidar data and the noise points \cite{li2020deep}.

\subsubsection{Multi-modal Road Extraction}
As analyzed above, each modality has individual benefits and drawbacks, so it's wise to aggregate their complementary information for extracting traffic roads effectively. In literature, numerous methods have been proposed to identify road areas using both aerial images and Lidar data, because of the accessibility of these data. For instance, Hu \textit{et al.} \cite{hu2004automatic} first segmented the primitives of roads from both optical images and Lidar data, and then detected road stripes with an iterative Hough transform algorithm to form the final road network by topology analysis. Parajuli \textit{et al.} \cite{parajuli2018fusion} developed a modular deep convolution network called TriSeg, in which two SegNet \cite{badrinarayanan2017segnet} were used to extract features respectively from aerial images and Lidar data, and another SegNet fused modular features to estimate the final road maps.
However, neither of aerial images and Lidar data can provide sufficient information to discover the traffic roads heavily occluded by trees, thus some recent works incorporated aerial images and vehicle trajectories to identify road areas. For instance, Sun et al. \cite{sun2019leveraging} fed the concatenation of rendered trajectory heat-maps and aerial images into different backbone networks (e.g., UNet \cite{ronneberger2015u}, Res-UNet \cite{zhang2018road}, LinkNet \cite{chaurasia2017linknet} and D-LinkNet~\cite{zhou2018d}) to estimate those traffic roads. In \cite{wu2020deepdualmapper}, trajectory maps and aerial images were first fed into different networks respectively for feature extraction, and then the modular features at different layers were fused to predict the final roads. Despite progress, such a concatenation or fusion manner can not fully exploit the complementarities of different modalities, and more effective methods are desired for multimodal road extraction.

\subsection{Message Passing Mechanism}
In the field of machine learning, message passing \cite{ross2011learning,winn2005variational} refers to information interactions between different entities. A large number of works have shown that such a mechanism can effectively facilitate deep representation learning. For instance, Wang \textit{et al.} \cite{wang2018dividing} introduced an inter-view message passing module to enhance the view-specific features for action recognition, while Liu \textit{et al.} \cite{liu2019crowd} propagated information among multiscale features to model the scale variations of people. In graph convolution networks, the message passing mechanism is usually embedded to aggregate information from neighboring nodes \cite{zhang2020dynamic,zhong2020hierarchical,liu2020physicalvirtual,9206045,chen2021cross}. Recently, this mechanism has also been adopted for cross-modal representation learning. For instance, Wang \textit{et al.} \cite{wang2019camp} addressed the text-image retrieval problem by transferring multimodal features and computing their matching scores. Nevertheless, most of these previous methods propagated information in a local manner (e.g., at short range). Without capturing global information, these methods may fail to discover the occluded roads that meanwhile have very few trajectories, as shown in Fig. \ref{fig:challenge}-(d). Therefore, more effective approaches are desired to fully exploit the complementary information of aerial images and crowdsourced trajectories for traffic road extraction.

\begin{figure*}[t]
  \begin{center}
     \includegraphics[width=1.85\columnwidth]{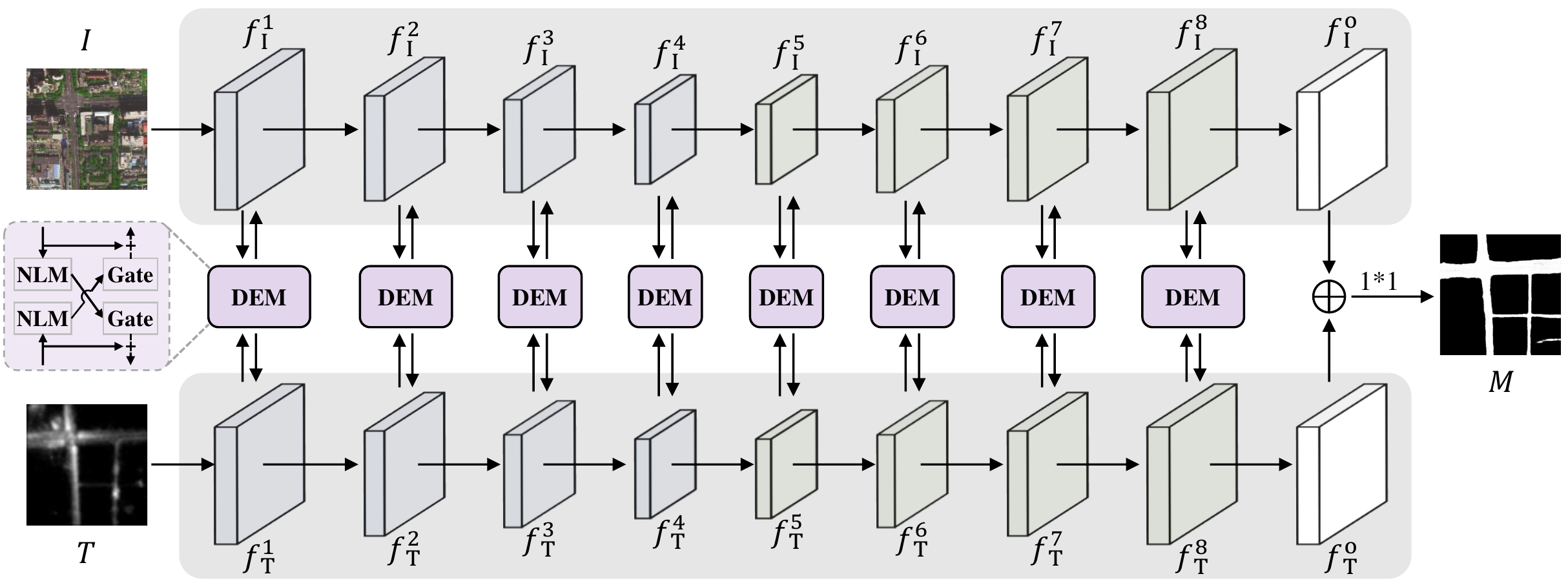}
  \vspace{-4mm}
  \end{center}
   \caption{The architecture of the proposed Cross-Modal Message Propagation Network (CMMPNet) for multimodal road extraction. Specifically, our CMMPNet is composed of {\bf\color{red}(i)} two deep AutoEncoders that take an aerial image and a trajectory heat-map respectively to learn modality-specific features, and {\bf\color{red}(ii)} a Dual Enhancement Module (DEM) that dynamically propagates the non-local messages (NLM, i.e, local one and global one) of every modality with gated functions to enhance the representation of another modality. The final features of the image and trajectory heat-map are concatenated to generate a traffic road map.}
\vspace{0mm}
\label{fig:network-structure}
\end{figure*}

\section{Preliminaries}\label{sec:preliminary}

In this section, we first introduce how to generate trajectory heat-maps from raw GPS data and then formally define the problem of image+trajectory based road extraction.

\textbf{Raw Trajectory Samples}: With the rapid growth of vehicle ownership, we can easily collect a mass of vehicles' GPS trajectories to construct a large-scale trajectory database \cite{liu2019contextualized,lou2020probabilistic,liu2020dynamic}. In this database, each trajectory sample can be represented as a tuple $\{{vid}, {lon}, {lat}, {t}, {sp}, {si}\}$, where ${vid}$ is the ID of a vehicle, $lon$ and $lat$ are the longitude and latitude at timestamp ${t}$. Term $sp$ denotes the vehicle's speed. $si$ is the trajectory sampling interval and different vehicles have different sampling settings. We notice that some early works \cite{liu2012mining} manually generated some virtual samples on the line segment between two consecutive real samples to augment the trajectory quantity. Nevertheless, this would cause a lot of noise in complex scenarios, since the real-world vehicles may have large sampling intervals (such as ${si}$ is mainly set to 10, 60, 180, and 300 seconds in Beijing~\cite{sun2019leveraging}) and it is difficult to accurately infer the virtual trajectories under these settings. Thanks to the crowdsourcing mechanism, adequate trajectories can be easily collected nowadays. Therefore, we only use the real trajectory samples in this work.

\textbf{Trajectory Heat-map Generation}:
For deep neural networks, matrix or tensor is one of the most common formats of input. Thus we need to transform the raw GPS data into 2D trajectory heat-maps before feeding them into networks. The whole transformation process is shown in Fig.~\ref{fig:render_map}.
Specifically, give an aerial image with a resolution $H{\times}W$, we first search out all trajectory samples in its coordinate range $[{lon}_l, {lat}_l]$*$[{lon}_u, {lat}_u]$, where the subscripts $l$ and $u$ denote the lower and upper bounds, respectively. These samples are then projected into a $H{\times}W$ greyscale map by counting the number of samples projected at every pixel. In this map, the pixels of road areas usually have high values, while the pixel values in non-road regions are very small, even zero. This would facilitate the discovery of traffic roads.
However, we find that such a projected map has two minor defects. First, some infrequently-traveled roads are so pale that they are hard to be recognized. Second, this map is too coarse and sharp. For example, two adjacent pixels in road areas may have values of different scales, or even a road pixel does not match any projected samples.
Inspired by Kernel Density Estimation frequently used for trajectory processing \cite{liu2012mining}, we normalize the projected map with a logarithm function and apply a $3{\times}3$ Gaussian kernel filter for smoothing. The involved Gaussian filter can also eliminate trajectory noises to a certain degree \cite{Gaussian}. In this way, the final trajectory heat-map becomes smooth and traffic roads are more distinct from backgrounds.

\textbf{Image+Trajectory based Road Extraction}: Given a $H{\times}W$ aerial image $I$ and the corresponding trajectory heat-map $T$, our goal is to automatically predict a $H{\times}W$ binary road map
\begin{equation}
M=\mathcal{F}(\{I,T\}, \theta),
\end{equation}
where $\mathcal{F}(\cdot)$ is a mapping function with learnable parameters $\theta$. Specifically, the pixels within road areas are supposed to have high response value (i.e.. 1), while the response values of background pixels should be 0.

\section{Methodology}\label{sec:CMMP}
\subsection{Framework Overview}
As mentioned above, aerial images and vehicle crowdsourced trajectories are complementary for traffic road extraction. To recognize unconstrained roads effectively and robustly, we propose a Cross-Modal Message Propagation Network (CMMPNet), which mutually enhances the hierarchical features of different modalities for better capturing their complementary information. As shown in Fig.~\ref{fig:network-structure}, our CMMPNet is composed of {\bf\color{red}(i)} two deep AutoEncoders for modality-specific feature learning and {\bf\color{red}(ii)} a Dual Enhancement Module (DEM) for cross-modal feature refinement. In this subsection, we mainly introduce the architecture of CMMPNet, whose specific components are described in the following subsections.

Specifically, given an aerial imagery $I$ and a trajectory heat-map $T$ with a resolution ${H{\times}W}$, we first explicitly learn modality-specific representations by feeding them into different AutoEncoders, each of which consists of four encoding blocks and four decoding blocks. As shown in Fig.~\ref{fig:network-structure}, the first AutoEncoder takes $I$ as input and extracts a group of image features:
\begin{equation}
f_I = \{f_{I}^{1}, f_I^{2}, f_I^{3}, f_I^{4}, f_I^{5}, f_I^{6}, f_I^{7}, f_I^{8}\},
\end{equation}
where the first four features are the outputs of encoding blocks and the remaining four features are the output of decoding blocks.
With the same architecture, the second AutoEncoder extracts a group of trajectory features:
\begin{equation}
f_T = \{f_{T}^{1}, f_T^{2}, f_T^{3}, f_T^{4}, f_T^{5}, f_T^{6}, f_T^{7}, f_T^{8}\},
\end{equation}
from the input trajectory heat-map $T$.

Rather than directly fuse image and trajectory features with concatenation \cite{li2019fusing} or weighted addition \cite{wu2020deepdualmapper}, we fully capture the multimodal complementary information through enhancing their features mutually with a message passing mechanism.
For each pair of multimodal feature $\{f_{I}^{i}, f_{T}^{i}\}$, we employ the proposed DEM to generate two enhanced features $\{\widehat{f}_{I}^{i}, \widehat{f}_{T}^{i}\}$ with their complementary information. This process can be formulated as:
\begin{equation}
\widehat{f}_{I}^{i}, \widehat{f}_{T}^{i} = \text{DEM}(f_{I}^{i}, f_T^{i}), {~~~} i=1,2,...,8.
\end{equation}
These enhanced features are then fed into the next block of individual AutoEncoder respectively for further representation learning. For convenience, the final outputs of image AutoEncoder and trajectory AutoEncoder are denoted as $f_I^o$ and $f_T^o$, and they have the same resolution ${H{\times}W}$. Here, $f_I^o$ and $f_T^o$ are jointly utilized to predict a probability map $M \in R^{H{\times}W}$ for traffic roads with the following formulation:
\begin{equation}
M = Conv(f_{I}^{o}{\oplus}f_T^{o}, \mathbb{W}_{1*1}),
\end{equation}
where $\oplus$ denotes feature concatenation and $\mathbb{W}_{1*1}$ refers to the parameters of a 1*1 convolutional layer. For each position $(x, y)$, it can be regarded as a road region only when $M(x, y)$ is greater than a given threshold.

It's worth noting that our method is universal for multi-modal road extraction. Except for image+trajectory data, the proposed CMMPNet can also be directly employed to recognize traffic roads with image+lidar data. The university of our method would be verified in Section \ref{sec:experiment} and \ref{sec:lidar_experiment}.

\subsection{Modality-Specific Feature Learning}
In the previous work \cite{sun2019leveraging}, aerial images and trajectory heat-maps were directly concatenated to feed into the same network, which caused that their features were over mixed and their complementarities were missed to some extent.
To address this problem, we feed the given aerial image and the corresponding trajectory heat-map into different networks to learn modality-specific features. Optimized with individual parameters, these features well preserve the specific information of each modality, thus can be further utilized for mutual refinement.

\begin{table}
  \caption{The configuration of our AutoEncoder. In the first convolutional layer, the input channel $C_i$ is set to 3 for aerial images and 1 for trajectory heat-maps, and the stride is set to 2. In each block, DR denotes the downsampling ratio of resolution and $C_o$ is the channel number of output. MP denotes a $2\times2$ max-pooling layer. \textit{Res}, \textit{Up} and \textit{Inter} refer to the residual unit, upsampling unit and interim unit described in Fig. \ref{fig:network-block}.}
  \vspace{-2mm}
  \centering
  \resizebox{8.95cm}{!} {
    \begin{tabular}{c|c|c|c|c}
    \hline
    \multirow{2}{*}{\textbf{Block}} & \multirow{2}{*}{\textbf{Configuration}} & \multicolumn{3}{c}{\textbf{Output}}  \\
    \cline{3-5}
     & & \textbf{Sign} & \textbf{DR} & \textbf{$C_o$}\\
    \hline\hline
    - & Conv(7,$C_i$,64,s=2) &   & 1/2 & 64  \\
    encoding-1 & MP$\rightarrow$3*Res(64, 64)  & $f^{1}$ & 1/4 & 64\\
    encoding-2 & MP$\rightarrow$Conv({~~}64,128)$\rightarrow$3*Res(128,128) & $f^{2}$ & 1/8 & 128\\
    encoding-3 & MP$\rightarrow$Conv(128,256)$\rightarrow$5*Res(256,256) & $f^{3}$ & 1/16 & 256 \\
    encoding-4 & MP$\rightarrow$Conv(256,512)$\rightarrow$2*Res(512,512) & $f^{4}$ & 1/32 & 512 \\
    \hline
    - & Inter(512,512) &   & 1/32 & 512  \\
    \hline
    decoding 1 & Up(512,256) + $f_{3}$ & $f^{5}$ & 1/16 & 256  \\
    decoding 2 & Up(256,128) + $f_{2}$ & $f^{6}$ & 1/8 & 128   \\
    decoding 3 & Up(128,{~~}64) + $f_{1}$ & $f^{7}$ & 1/4 & 64   \\
    decoding 4 & Up({~}64,{~~}64){~~~~~~~} & $f^{8}$ & 1/2 & 64 \\
    - & TConv(4,64,32,2)$\rightarrow$Conv(3,32,32) & $f^o$ & 1 & 32   \\
    \hline
    \end{tabular}
  }
  \vspace{2mm}
  \label{tab:network_configuration}
\end{table}

\begin{figure}
\centerline{
\includegraphics[width=1\columnwidth]{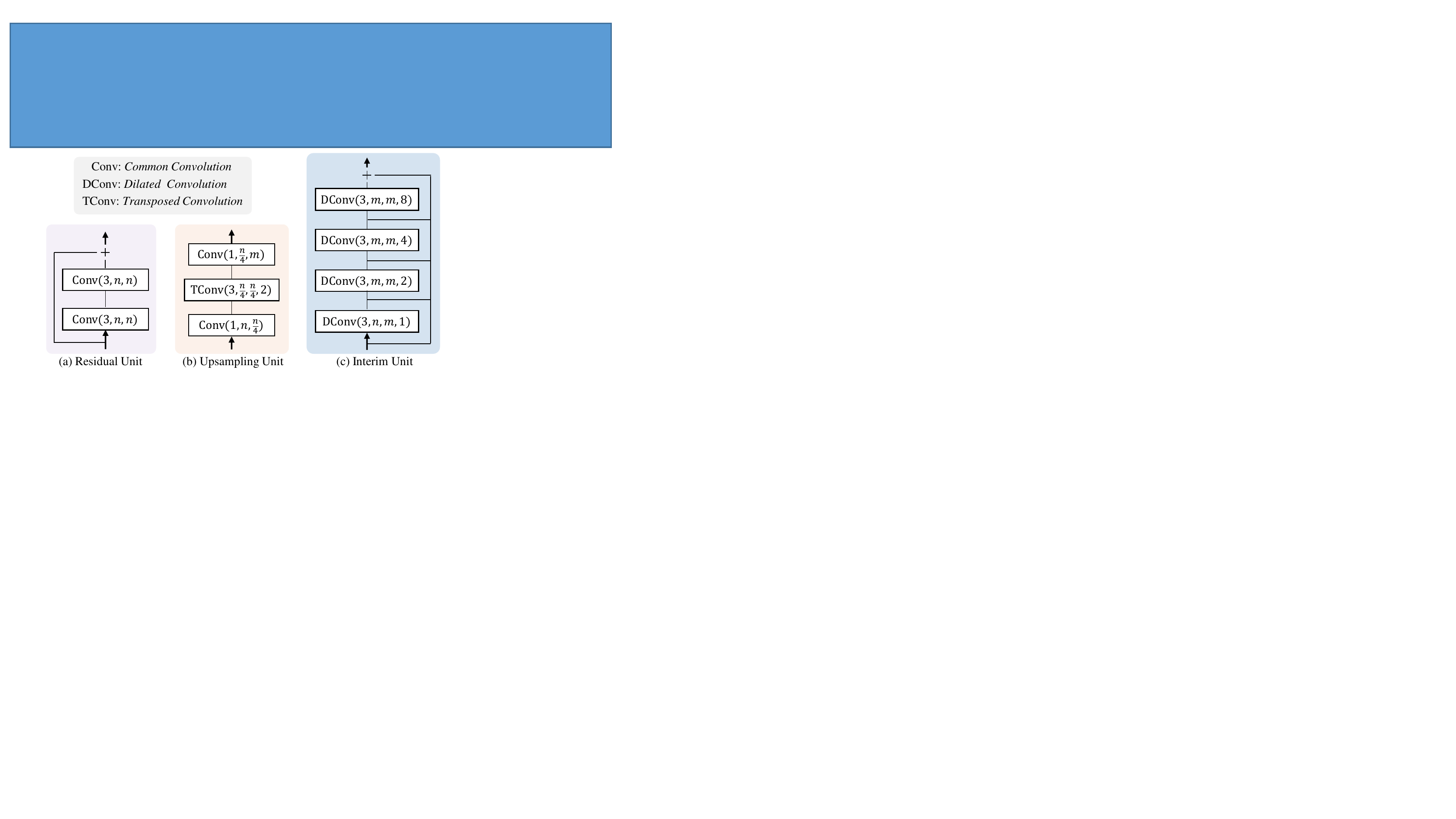}
}
\vspace{-3mm}
   \caption{The architecture of Residual Unit, Upsampling Unit, and Interim Unit. $Conv(k, n, m)$ denotes a $k{\times}k$ standard convolution, whose input channel is $n$ and output channel is $m$. $DConv(k, n, m, r)$ refers to a dilated convolution with a dilated ratio $r$ and $TConv(k, n, m, s)$ is a transposed convolution with a stride $s$.}
\vspace{0mm}
\label{fig:network-block}
\end{figure}

To maintain the high resolution of final outputs, two AutoEncoders are adopted intentionally to extract modality-specific features. Notice that various AutoEncoders (e.g, Res-UNet \cite{zhang2018road}, LinkNet \cite{chaurasia2017linknet}, and D-LinkNet \cite{zhou2018d}) are suitable to serve as the backbone network of our framework. Since these networks have similar architectures, we take D-LinkNet based AutoEncoder as an example to demonstrate the details of modality-specific feature learning. As shown in Table \ref{tab:network_configuration}, both the image AutoEncoder and trajectory AutoEncoder are mainly composed of four encoding blocks and four decoding blocks. Specifically, we first use a convolutional layer to extract initial features and then feed them into the following four encoding blocks, each of which consists of a $2\times2$ max-pooling layer and multiple residual units. As shown in Fig. \ref{fig:network-block}-(a), each residual unit contains two $3\times3$ convolutional layers and a skip layer.
After the encoding stage, an interim unit is adopted to capture more spatial context by expanding the receptive field with four dilated convolutional layers. At the decoding stage, four decoding blocks are utilized to enlarge the resolutions of features progressively. Specifically, each decoding block is developed as an upsampling unit, which consists of two convolutional layers for channel adjustment and a transposed convolutional layer for feature upsampling, as shown in Fig. \ref{fig:network-block}-(c). To simultaneously exploit the lower-level information and high-level information, we incorporate the features of encoding blocks and decoding blocks with element-wise addition. Finally, we fully restore the resolution of feature to $H{\times}W$ with a transposed convolution and apply a $3\times3$ convolutional layer to generate the final modality-specific feature $f^o \in R^{H{\times}W{\times}32}$. Note that our image AutoEncoder and trajectory AutoEncoder have individual parameters, thus they can effectively capture and preserve the specific information of each modality.

\subsection{Cross-modal Feature Refinement}\label{sec:DEM}
After modality-specific feature learning, we refine these features mutually with a Dual Enhancement Module (DEM) based on the message passing mechanism. In this module, a Non-Local Message Propagator and a Gated Message Propagator are integrated to dynamically transmit the local and global message of each unimodal feature to complement the feature of another modality. Absorbing the complementary information of other modalities, each unimodal feature becomes more reasonable and robust. In this subsection, we take the refinement of features $f^i_I$ and $f^i_T \in R^{h{\times}w{\times}c}$ as an example to demonstrate the working mechanism of the tailor-designed DEM. Note that $h$, $w$, and $c$ are the height, width, and channel number of these features.

\begin{figure}
\centerline{
\includegraphics[width=0.90\columnwidth]{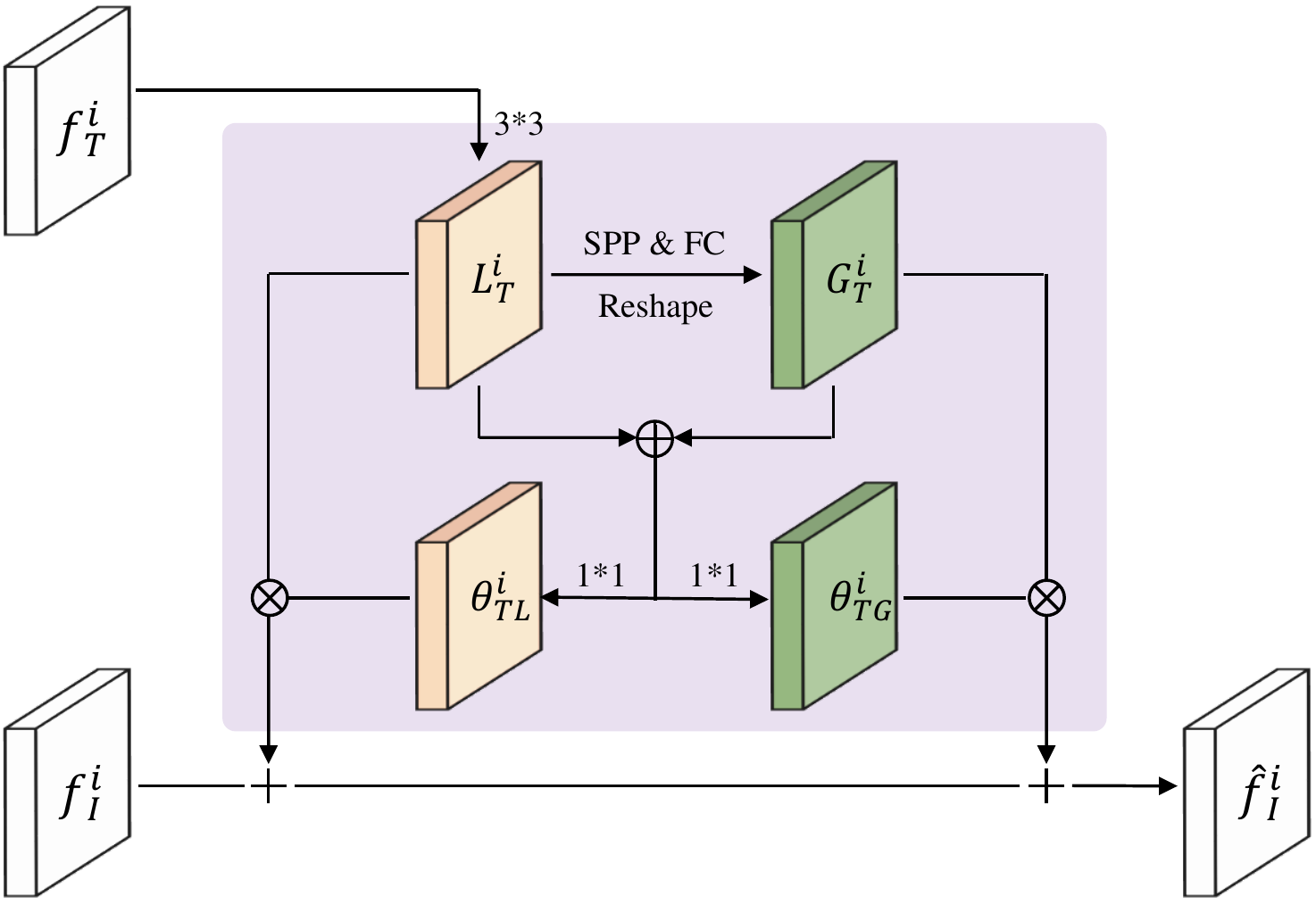}
}
\vspace{-3mm}
   \caption{The architecture of Dual Enhancement Module. This figure mainly illustrates how to enhance the image feature $f^i_I$ with the information extracted from the trajectory feature $f^i_T$. The cross-modal information from $f^i_T$ to $f^i_I$ is obtained by dynamically fusing the local information $L^i_T$ and global information $G^i_T$ with the learnable fused weights ${\theta}^i_{TL}$ and ${\theta}^i_{TG}$. This architecture can also be employed to enhance $f^i_T$. SPP and FC are the abbreviations of Spatial Pyramid Pooling~\cite{he2015spatial} and Fully-Connected layer, respectively. + and $\otimes$ denote the element-wise addition and multiplication, and $\oplus$ refers to feature concatenation.}
\vspace{-1mm}
\label{fig:DEM}
\end{figure}

\subsubsection{Non-Local Message Propagator}
Unlike previous works \cite{zhang2018bi,liu2019crowd} that only used local cues, our method explores both the local and global information for feature enhancement. Here we mainly introduce how to utilize the information of trajectory feature $f^i_T$ to enhance the image feature $f^i_I$. The refinement of $f^i_T$ is performed with the same process.

As shown in Fig. \ref{fig:DEM}, we first extract a local information map $L^i_T \in R^{h{\times}w{\times}c}$ by feeding $f^i_T$ into a $3\times3$ convolutional layer. Then we aggregate the local information at different locations to generate a global information map. Rather than use the compute-intensive non-local module proposed in \cite{wang2018non}, we employ a lightweight $N$-level Spatial Pyramid Pooling (SPP~\cite{he2015spatial}) and a Fully-Connected (FC) layer for global information generation. Specifically, at the $i$-th level ($i$=1,2...,$N$), $L^i_T$ is divided into $2^{i-1}\times2^{i-1}$ regions, each of which has a dimension of $\frac{h}{2^{i-1}}{\times}\frac{w}{2^{i-1}}{\times}c$ and is fed into a $\frac{h}{2^{i-1}}{\times}\frac{w}{2^{i-1}}$ max-pooling layer to obtain a $1{\times}1{\times}c$ information vector. Further, the information vectors at all levels are concatenated and fed into the FC layer with $c$ output neurons to generate a global information vector. This global vector is copied $h{\times}w$ times and reshaped to form the global information map $G^i_T \in R^{h{\times}w{\times}c}$.
After obtaining the local map $L^i_T$ and global map $G^i_T$, we can easily propagate them to refine $f_I^{i}$ with the following formulation:
\begin{equation}
\widehat{f}^i_I = f^i_I + L^i_T + G^i_T,
\label{I_refine}
\end{equation}
where $\widehat{f}^i_I$ is the enhanced image feature and the operator ``+'' denotes a element-wise addition. In the same way, we can compute the enhanced trajectory feature $\widehat{f}^i_T$ as follow:
\begin{equation}
\widehat{f}^i_T = f^i_T + L^i_I + G^i_I,
\label{T_refine}
\end{equation}
where $L^i_I$ and $G^i_I$ are the extracted local information map and global information map of image feature $f^i_I$.

\subsubsection{Gated Message Propagator}
In the previous propagator, the local and global information is transmitted statically, which isn't optimal for cross-modal refinement and even has disturbing effects at some locations. To alleviate this issue, a Gated Message Propagator is introduced to adaptively determine and propagate the complementary information. With multiple learnable gate functions, the beneficial information is transmitted and the disturbing information (e.g., visual cues of train tracks and the noises of trajectories)  is suppressed.

Specifically, we first introduce the computation of the gated weights of different information. As shown in Fig.~\ref{fig:DEM}, the trajectory information $L^i_T$ and $G^i_T$ is concatenated and fed into two $1\times1$ convolutional layers:
\begin{equation}
\begin{split}
{\theta}^i_{TL} &= \textit{Sigm}(\textit{Conv}(L^i_T \oplus G^i_T , \mathbb{W}^i_{TL})), \\
{\theta}^i_{TG} &= \textit{Sigm}(\textit{Conv}(L^i_T \oplus G^i_T , \mathbb{W}^i_{TG})),
\end{split}
\end{equation}
where ${\theta}^i_{TL}, {\theta}^i_{TG} \in R^{h{\times}w{\times}c}$ are the gated weights of $L^i_T$ and $G^i_T$, respectively. $\mathbb{W}^i_{TL}$ and $\mathbb{W}^i_{TG}$ are the parameters of convolutional layers, and $\textit{Sigm}()$ is an element-wise sigmoid function. In this same way, we can also compute the gated weights ${\theta}^i_{IL}, {\theta}^i_{IG} \in R^{h{\times}w{\times}c}$ for the information $L^i_I$ and $G^i_I$.
Finally, we re-weight each information with individual gated weight and then preform the dynamic message propagation. Therefore, Eq. \ref{I_refine} and \ref{T_refine} have become:
\begin{equation}
\begin{split}
\hat{f}_{I}^{i} &= f_I^{i} + {\theta}^i_{TL} \otimes L^i_T + {\theta}^i_{TG} \otimes G^i_T, \\
\hat{f}_{T}^{i} &= f_T^{i} + {\theta}^i_{IL} \otimes L^i_I {~} + {\theta}^i_{IG} \otimes G^i_I,
\end{split}
\end{equation}
where $\otimes$ denotes an element-wise multiplication.

\subsection{Implementation Details}\label{sec:details}
In this work, we implement the proposed CMMPNet on the representative deep learning platform PyTorch~\cite{paszke2019pytorch}. First, we perform data augmentation to alleviate the overfitting issue. Specifically, all training samples including the satellite images, trajectory heat-maps and ground-truth maps are {\bf{i)}} flipped horizontally or vertically, {\bf{ii)}} rotated by 90, 180, 270 degrees, {\bf{iii)}} randomly cropped with a size range of [0.7, 0.9] and resized to the original resolution. After augmentation, the number of training samples is enlarged by 7 times. We then determine the hyper-parameters of our framework. The filter weights of all convolutional layers and Fully-Connected layers are uniformly initialized by Xavier~\cite{glorot2010understanding}. The batch size is set to 4 and the learning rate is set to 0.0002. Finally, we apply the Adam~\cite{kingma2014adam} optimizer to train our CMMPNet for 30 epochs by minimizing the Binary Cross-Entropy Loss between the generated road maps and the corresponding ground-truth maps.

\section{Experiments}\label{sec:experiment}
In this section, we first introduce the experiment settings of image+trajectory based road extraction. We then compare the proposed CMMPNet with existing state-of-the-art approaches and finally conduct extensive ablation studies to verify the effectiveness of each component in our network.

\subsection{Settings}
{\bf{Datasets:}} In this work, our experiments are mainly conducted on the BJRoad dataset \cite{sun2019leveraging}, which is captured in Beijing, China. Specifically, this benchmark consists of 350 high-resolution aerial images that cover a large geographic area of about 100 square kilometers and around 50 million trajectory records of 28 thousand vehicles. The resolution of aerial images is $1024\times1024$ and each pixel denotes a 0.5m$\times$0.5m region in the real world. For each aerial image, a $1024\times1024$ trajectory heat-map is generated with the preprocessing described in Section \ref{sec:preliminary}, and the corresponding ground-truth (GT) map is manually created by masking out the pixel of traffic roads. Finally, this dataset is officially divided into three partitions: 70\% samples are adopted for training, 10\% for validation, and the rest 20\% for testing.

Following the previous work \cite{wu2020deepdualmapper}, we also perform experiments on the Porto dataset, which covers a geographic area of about 209 square kilometers in Porto, Portugal. This dataset contains a mass of crowdsourced trajectories generated by 442 taxis from 2013 to 2014. On this dataset, we adopt a five-fold cross-validation setting, since the details of training/testing sets are not provided in \cite{wu2020deepdualmapper}. Specifically, the aerial image of the whole area is first cut into 6,048 non-overlapping sub-images with a resolution of $512\times512$. These sub-images are then randomly divided into five equal parts. For the $i$-th validation, the $i$-th part is used for testing, and the remaining parts are used for training. Finally, the mean and variance of five validations are reported.

{\bf{Evaluation Details:}}
Given a probability map $M$, we need to determine an estimated road map $M_e \in R^{H{\times}W}$ before evaluation. Same to \cite{sun2019leveraging}, a pixel $(x,y)$ is predicted as a road region in our work, if the response value of $M(x,y)$ is greater than 0.5. Following previous works \cite{long2015fully,zhou2018d}, we adopt Intersection over Union\footnote{\url{https://en.wikipedia.org/wiki/Jaccard_index}} (IoU) to evaluate the performance for road extraction. Specifically, the IoU score between an estimated map $M_e$ and its corresponding ground-truth map $M_g$ is computed by:
\begin{equation}
IoU(M_e, M_g) = {\frac{|M_e{\cap}M_g|}{|M_e{\cup}M_g|}},
\end{equation}
where $|M_e{\cap}M_g|$ denotes the pixel number in the intersection set of $M_e$ and $M_g$, and $|M_e{\cup}M_g|$ is the pixel number in their union set. There are two manners for computing the IoU of all testing samples. The first manner is to compute the IoU of each sample and then average the IoU of all samples. Such a metric is termed as average IoU (A\_IoU). The second manner is to stitch the estimated maps of all samples into a global map and then compute an IoU score. This metric is termed as global IoU (G\_IoU). Since different IoU metrics were used in previous works, we would report the results of both A\_IoU and G\_IoU in the following sections.

\begin{table}[t]
  \caption{The performance of different methods on the testing set of BJRoad dataset. Our CMMPNet outperforms all existing approaches with large margins.}
  \vspace{-2mm}
\newcommand{\tabincell}[2]{\begin{tabular}{@{}#1@{}}#2\end{tabular}}
  \centering
    \begin{tabular}{c|c|c}
    \hline
    \multicolumn{1}{c|}{\textbf{Method}} & \textbf{A\_IoU (\%)} & \textbf{G\_IoU (\%)} \\
    \hline\hline
      {~~}DeepLab (v3+)~\cite{chen2018encoder} &  50.81 & - \\
      {~~~~~~~~~~~~~~}UNet~\cite{ronneberger2015u} &  54.88 & - \\
      {~~~~~~~~~}Res-UNet~\cite{zhang2018road} & 54.24 & - \\
      {~~~~~~~~~~~}LinkNet~\cite{chaurasia2017linknet} &  57.89 & - \\
      {~~~~~~~~}D-LinkNet~\cite{zhou2018d} & 57.96 & - \\
      {~~~~~~~~~}Sun et al.~\cite{sun2019leveraging} &  59.18 & - \\
      DeepDualMapper \cite{wu2020deepdualmapper} & 60.91 & 61.54 \\
    \hline
    {~~}Res-UNet+CMMPNet    &  {\bf62.58} & {\bf63.03} \\
      {~~~~}LinkNet+CMMPNet &  {\bf63.09} & {\bf63.46} \\
      D-LinkNet+CMMPNet     &  {\bf62.85} & {\bf63.39} \\
    \hline
    \end{tabular}
  \label{tab:BJRoad_IoU}
\end{table}

\subsection{Comparison with State-of-the-Art Methods}
In this section, we compare our CMMPNet with seven deep learning-based approaches, including DeepLab (v3+) \cite{chen2018encoder}, UNet \cite{ronneberger2015u}, Res-UNet \cite{zhang2018road}, LinkNet \cite{chaurasia2017linknet}, D-LinkNet \cite{zhou2018d}, Sun \textit{et al.} \cite{sun2019leveraging}, and DeepDualMapper \cite{wu2020deepdualmapper}. Specifically, these compared methods are reimplemented for multimodal road extraction. In particular, DeepDualMapper feeds aerial images and trajectory heat-maps into different backbone networks\footnote{In DeepDualMapper, the original backbone network is UNet. However, our reimplemented DeepDualMapper based on UNet performs poorly. Thus in this work, we adopt D-LinkNet as the backbone to reimplement DeepDualMapper and this model can obtain competitive performance on different datasets.} and then fuses their features with a gated fusion module, while other methods directly take the concatenation of aerial images and trajectory heat-maps as input. Moreover, all the compared methods except DeepLab (v3+) and DeepDualMapper are equipped with 1D transpose convolution to better model traffic roads \cite{sun2019leveraging}. Notice that the first six compared methods were implemented by \cite{sun2019leveraging}, and we utilize the official code of \cite{sun2019leveraging} to implement DeepDualMapper and our method with the same data partition. As mentioned above, our CMMPNet can be developed with various AutoEncoders. We hence evaluate multiple implementations of CMMPNet based on different AutoEncoders, such as Res-UNet, LinkNet, and D-LinkNet.

\begin{table}[t]
  \caption{The performance of different methods on the PORTO dataset. Five-fold cross-validation is conducted on this dataset. The mean and variance of five validations are reported in this table.}
  \vspace{-2mm}
\newcommand{\tabincell}[2]{\begin{tabular}{@{}#1@{}}#2\end{tabular}}
  \centering
    \begin{tabular}{c|c|c}
    \hline
    \multicolumn{1}{c|}{\textbf{Method}} & \textbf{A\_IoU (\%)} & \textbf{G\_IoU (\%)} \\
    \hline\hline
       {~~~~~~~~}D-LinkNet~\cite{zhou2018d}          & 72.82$\pm$0.47 & 72.92$\pm$0.45 \\
      {~~~~~~~~~}Sun et al.~\cite{sun2019leveraging} & 72.94$\pm$0.71 & 73.04$\pm$0.63 \\
      DeepDualMapper \cite{wu2020deepdualmapper}     & 73.67$\pm$0.51 & 73.91$\pm$0.51 \\
    \hline
      D-LinkNet+CMMPNet                              & 74.56$\pm$0.46 & 74.66$\pm$0.41 \\
    \hline
    \end{tabular}
  \label{tab:Porto_IoU}
\end{table}

\begin{figure*}
\centerline{
\includegraphics[width=1.7\columnwidth]{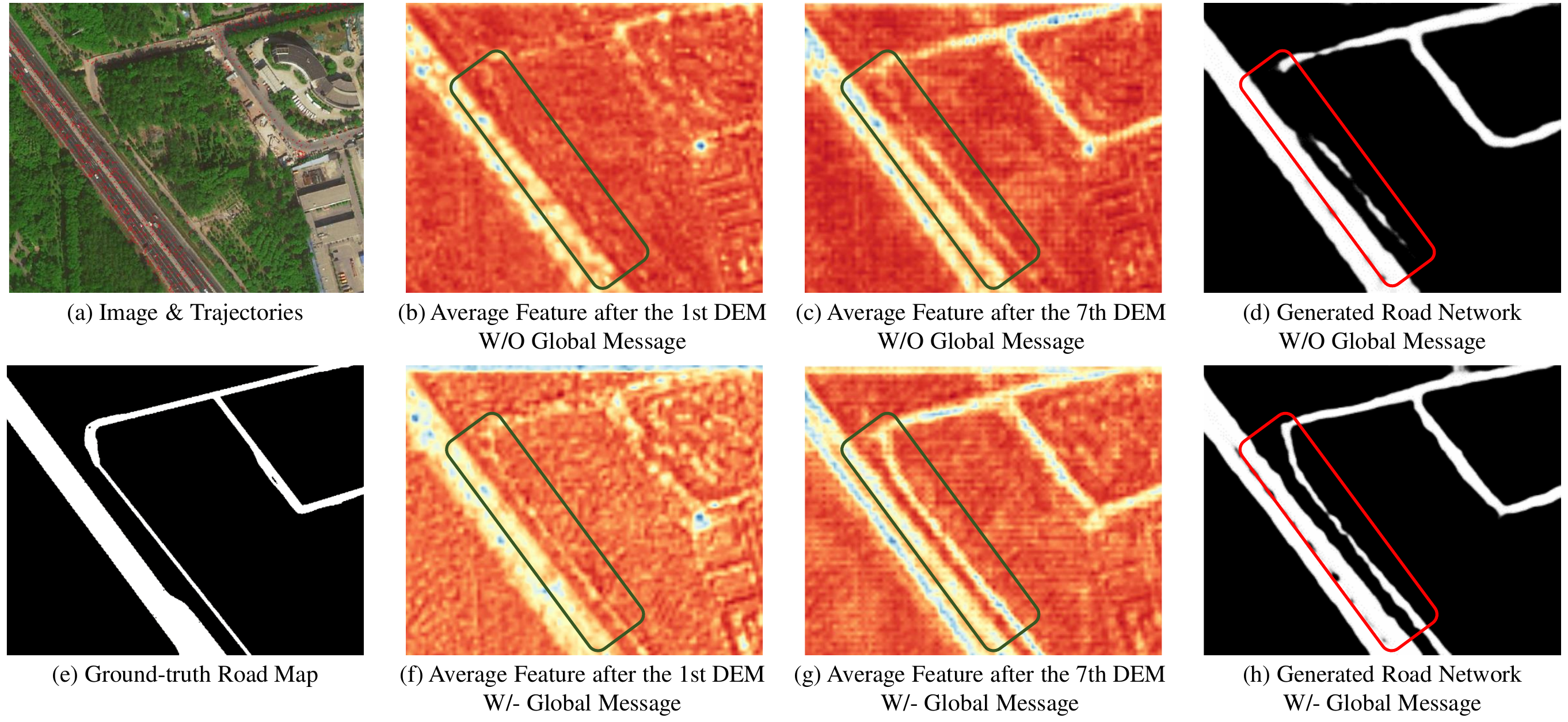}
}
\vspace{-3mm}
   \caption{Visualization of the feature maps and traffic road maps generated with/without global information on the testing set of BJRoad dataset. (a) is the input aerial image with trajectory points, and (e) is the ground-truth road map. (b) and (c) are the average maps of image feature and trajectory feature after the 1st/7th DEM without the global message, while (d) is the generated road network without the global message. (f), (g) and (h) are the generated feature maps and the road network using both the local message and global message. We can observe that our method can generate more discriminative features and recognize the occluded/unimpressive traffic roads effectively when performing road extraction with global information.}
\vspace{0mm}
\label{fig:global_message}
\end{figure*}

The performance of all methods on the BJRoad dataset is summarized in Table \ref{tab:BJRoad_IoU}. We can observe that DeepLab (v3+) obtains the worst A\_IoU 50.81\% probably because of parameter overfitting. Sun \textit{et al.} \cite{sun2019leveraging} utilized various techniques (e.g., different sampling intervals and GPS augmentation) to obtain an improved A\_IoU 59.18\%. However, just directly feeding the concatenation of aerial images and trajectory heat-maps into networks, these methods have limited capabilities to capture the multimodal information, thus none of them can acquire an A\_IoU above 60\%. By fusing image and trajectory features with a gated fusion module, DeepDualMapper obtains a competitive A\_IoU 60.91\% and G\_IoU 61.54\%. Despite the progress, DeepDualMapper only use a fusion strategy rather than a mutual refinement strategy, thereby cannot address this task well.
In contrast, when learning modality-specific features explicitly and enhancing cross-modal features mutually, our method can fully exploit the complementary information of aerial images and crowdsourced trajectories. For this reason, the proposed CMMPNet outperforms all previous methods with large margins. For instance, Res-UNet+CMMPNet achieves a competitive A\_IoU 62.58\% and obtains a relative improvement of 15.37\%, compared with the original Res-UNet. By improving the A\_IoU from 57.96\% to 62.85\%, our D-LinkNet+CMMPNet also obtains a substantial improvement of 8.4\%, compared with the baseline D-LinkNet. Finally, with an impressive A\_IoU 63.09\% and a G\_IoU 63.46\%, our LinkNet+CMMPNet becomes the best-performing model. We notice that the performance of D-LinkNet+CMMPNet is slightly lower than that of LinkNet+CMMPNet. This is probably because D-LinkNet+CMMPNet contains two extra interim units and suffers from certain overfitting, although data augmentation has been performed.

Moreover, we compare the performance of our CMMPNet with three competitive models including D-LinkNet \cite{zhou2018d}, Sun et al.~\cite{sun2019leveraging}, and DeepDualMapper \cite{wu2020deepdualmapper} on the Porto dataset. As shown in Table \ref{tab:Porto_IoU}, all methods obtain much better results on this dataset, compared with their performance on the BJRoad dataset. The main reason is that the aerial images of Porto are clearer and the noises of trajectories are smaller \cite{wu2020deepdualmapper}. Despite the existing benchmarks are high, our CMMPNet still can boost the IOU with substantial margins, ranking first in performance on the Porto dataset. In summary, these comparisons greatly demonstrate the effectiveness of the proposed CMMPNet for image+trajectory based road extraction.

\subsection{Component Analysis}\label{sec:ablation}
After external comparison, we then perform extensive internal experiments to analyze the effectiveness of each module in the proposed CMMPNet. In this subsection, D-LinkNet is adopted as the backbone network and our implementation details have been described in Section \ref{sec:details}.

\begin{table}[t]
  \caption{The influence of Non-Local Message Propagator and Gated Message Propagator on the testing set of BJRoad dataset.}
  \vspace{-2mm}
\newcommand{\tabincell}[2]{\begin{tabular}{@{}#1@{}}#2\end{tabular}}
  \centering
    \begin{tabular}{ccc|cc}
    \hline
    \textbf{Local} & \textbf{Global} & \textbf{Gate} & \textbf{A\_IoU (\%)} & \textbf{G\_IoU (\%)} \\
    \hline\hline
    \checkmark & &            & 61.62 & 62.09 \\
    \checkmark & & \checkmark & 62.32 & 62.78 \\
    \hline
    \checkmark & \checkmark &            & 61.98 & 62.43 \\
    \checkmark & \checkmark & \checkmark & 62.85 & 63.39 \\
    \hline
    \end{tabular}
  \label{tab:global_gate}
\end{table}

{\bf{The effect of global message:}} In previous works \cite{lin2015deeply,teichmann2018convolutional}, local information is widely adopted, but global information is neglected. In this section, we implement several variants of CMMPNet to verify the effectiveness of global information. As shown in Table~\ref{tab:global_gate}, when propagating the global information extracted by SPP and FC layer, ``Local+Global'' model obtains an A\_IoU 61.98\% and a G\_IoU 62.43\%, and is better than ``Local'' model. With an A\_IoU 62.85\% and a G\_IoU 63.39\%, ``Local+Global+Gate'' model also outperforms ``Local+Gate'' model, whose A\_IoU is 62.32\% and G\_IoU is 62.78\%. Except for quantitative results, we also visualize some feature maps and traffic road maps generated by ``Local+Gate'' model and ``Local+Global+Gate'' model in Fig.~\ref{fig:global_message}. Note that those visualized feature maps are the channel-wise average of image features and trajectory features after DEM. We can observe that incorporating global information can generate more discriminative features and better recognize traffic roads, especially when the roads are occluded/unimpressive and the vehicle trajectories are rare in local regions. These quantitative and qualitative experiments show that global information is greatly effective for traffic road extraction.

{\bf{The effect of Gated Message Propagator:}} In this propagator, multiple gate functions are employed to dynamically propagate the complementary information. In this subsection, we also implement several variants to verify the effectiveness of this mechanism. As shown in Table~\ref{tab:global_gate}, after applying gate functions on ``Local'' model, A\_IoU increases from 61.62\% to 62.32\% and G\_IoU increases from 62.09\% to 62.78\%. Further, we can obtain a more substantial improvement (around 1\% on both A\_IoU and G\_IoU), when performing gate functions on ``Local+Global'' model. These comparisons show that this proposed propagator can facilitate robust road extraction using multimodal information.

\begin{figure*}[t]
  \begin{center}
     \includegraphics[width=2\columnwidth]{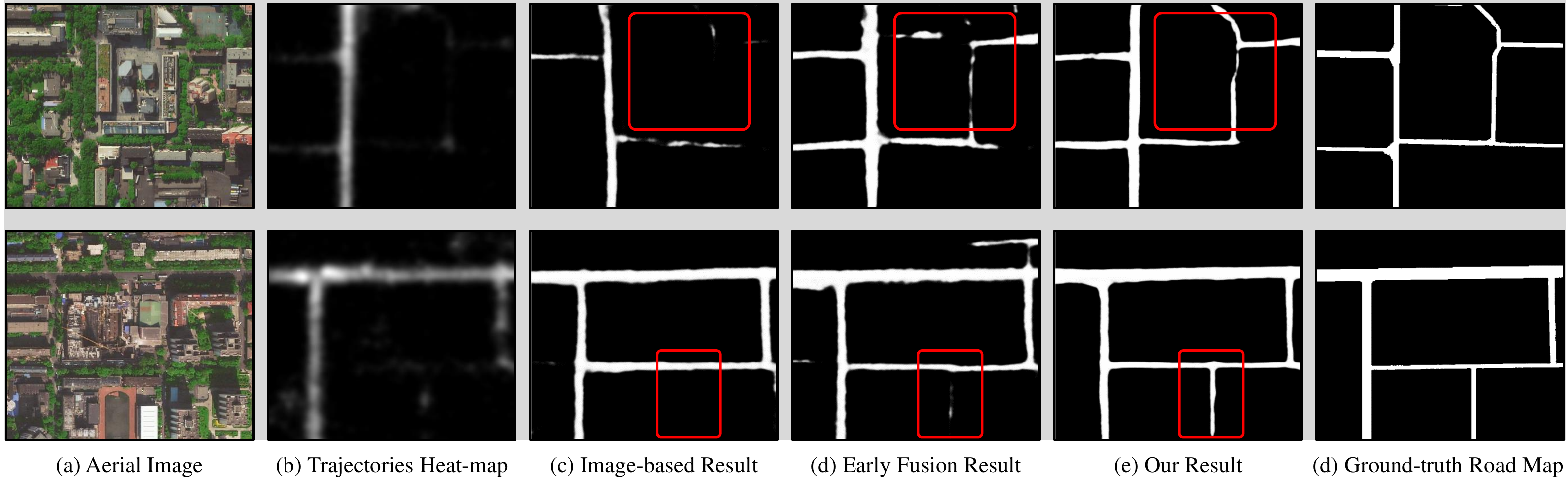}
  \vspace{-5mm}
  \end{center}
   \caption{Visualization of the traffic road networks generated by different methods on the testing set of BJRoad dataset. (a) and (b) are the input aerial images and trajectories heat-maps. (c) are the results that only aerial images are taken as input, while (d) are the results that the concatenation of images and heat-maps are taken as input. As shown in (e), the results of our CMMPNet are more accurate and are very similar to the ground-truth road networks.}
\vspace{0mm}
\label{fig:BJ_results}
\end{figure*}

\begin{figure}[t]
\centerline{
\includegraphics[width=0.95\columnwidth]{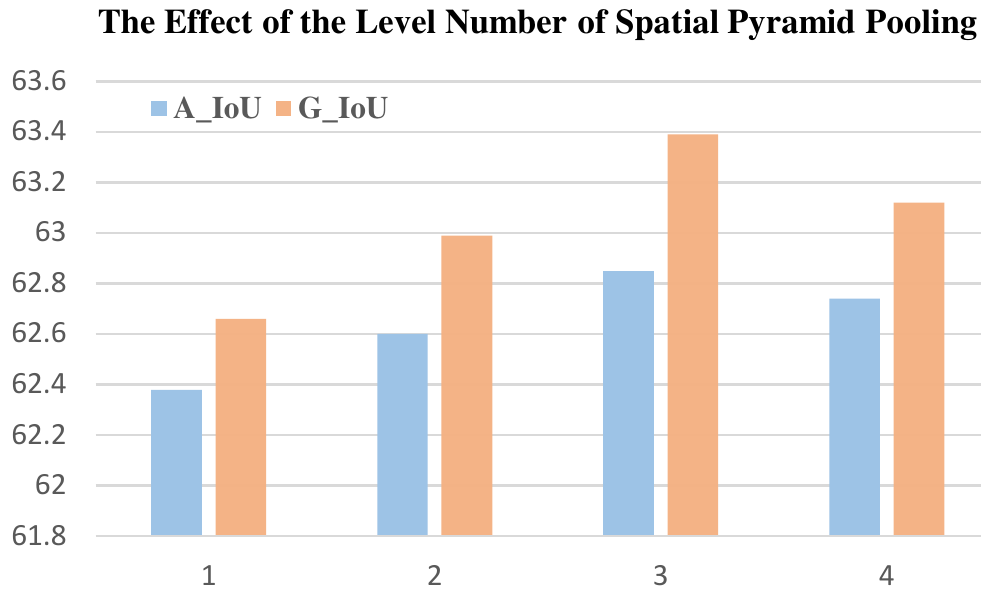}
}
\vspace{-4mm}
   \caption{The influence of the level number $N$ of Spatial Pyramid Pooling layer in Dual Enhancement Module on the testing set of BJRoad dataset. Our method achieves the best performance when $N$ is set to 3.}
\vspace{1mm}
\label{fig:level_number}
\end{figure}

{\bf{The configuration of Spatial Pyramid Pooling:}} In Non-Local Message Propagator, we employ a $N$-level SPP and an FC layer to extract global information. In this section, we explore the effect of the level number for road extraction using multimodal information. As shown in Fig.~\ref{fig:level_number}, when applying a global max-pooling ($N$=1), our CMMPNet obtains an A\_IoU 62.38\% and is slightly better than the ``Local+Gate'' model in Table \ref{tab:global_gate}, since global pooling can only provide some coarse and limited information. As the level number increases, the performance also gradually increases, and our method achieves the best A\_IoU 62.85\% and G\_IoU 63.39\% when $N$ is equal to 3. When $n$ increases to 4, the performance slightly drops, probably because of overfitting, i.e., the amount of parameters of the FC layer in DEM increases sharply as the level number increase. Therefore, the level number $N$ of Spatial Pyramid Pooling is uniformly set to 3 for road extraction.

\subsection{More Discussion}
{\bf{Unimodal Data vs. Multi-modal Data:}} We first explore whether multimodal data is reliably useful for traffic road extraction. As shown in Table~\ref{tab:multi_modal_manner}, when only feeding trajectory heat-maps into a D-LinkNet, we obtain a poor performance (A\_IoU 52.38\%, G\_IoU 52.90\%) on the BJRoad dataset. When only utilizing aerial images, we obtain an A\_IoU 59.79\% and a G\_IoU 60.24\%, which indicates that image data is more crucial than trajectory data. In contrast, when using the aerial images and trajectory heat-maps simultaneously, our CMMPNet and the early/late fusion models described in the next paragraph outperform the unimodal models consistently with an improvement of at least 1\% on IoU. This comparison demonstrates that multimodal data is more effective for traffic road extraction, because aerial images and vehicle crowdsourced trajectories have rich complementarities.

{\bf{Which multimodal learning manner is better?}} We then explore the effects of different multimodal learning manners. Except for the proposed CMMPNet, we also implement another two commonly-used manners, i.e., early fusion model and late fusion model. Specifically, the former feeds the concatenation of aerial images and trajectory heat-maps into a D-LinkNet. In the latter, aerial images and trajectory heat-maps are respectively fed into individual D-LinkNet, and their final features are concatenated to estimate the road maps. As shown in Table~\ref{tab:multi_modal_manner}, the early fusion model obtains an A\_IoU 61.11\% and a G\_IoU 61.53\%, slightly outperforming the late fusion model (A\_IoU 60.78\%, G\_IoU 61.24\%). This is because the multimodal information is utilized at different layers in the former, but just utilized once in the latter. Compared with these two models, our CMMPNet is more reasonable to learn modality-specific features and propagate cross-modal information hierarchically. For this reason, our method achieves an impressive A\_IoU 62.85\% and G\_IoU 63.39\%, and outperforms early/late fusion models with a large margin. This comparison shows the effectiveness of our CMMPNet for multimodal representation learning.

\begin{table}[t]
  \caption{The performance of different inputs and different representation learning manners on the testing set of BJRoad dataset.}
  \vspace{-2mm}
\newcommand{\tabincell}[2]{\begin{tabular}{@{}#1@{}}#2\end{tabular}}
  \centering
  \resizebox{0.36\textheight}{!}{%
    \begin{tabular}{c|c|c|c}
    \hline
    \multicolumn{1}{c|}{\textbf{Input}} & \textbf{Learning Manner} & \textbf{A\_IoU (\%)} & \textbf{G\_IoU (\%)} \\
    \hline\hline
      Trajectory  & - & 52.38 & 52.90  \\
      Image       & - & 59.79 & 60.24 \\
      \hline
      \multirow{3}{*}{Image+Trajectory}
       & {~~}Late Fusion & 60.78 & 61.24 \\
       & Early Fusion    & 61.11 & 61.53 \\
       & CMMPNet  & 62.85 & 63.39\\
    \hline
    \end{tabular}
  }
  \label{tab:multi_modal_manner}
\end{table}

{\bf{Significance of crowdsourced trajectories:}}
Although the IoU of aerial images is much better than that of trajectories, we argue that vehicle trajectories are crucial for the robustness of road extraction, especially when some cities (e.g, Chongqing and Chengdu, China) are greatly covered by fog and mist in aerial images. So here we explore to extract traffic roads from foggy images and crowdsourced trajectories. Since there are no foggy images in the BJRoad dataset, we need to generate some aerial images with heavy fog in advance. Specifically, for each cloudless image in BJRoad, we employ a fog effect renderer of Photoshop to generate a foggy image. After augmenting the training samples as described in Section~\ref{sec:details}, we reimplement the proposed CMMPNet and three other compared methods, including {\bf{(1)}} two unimodal models which feed foggy images or trajectory heat-maps into D-LinkNet, and {\bf{(2)}} an early fusion D-LinkNet model which takes the concatenation of foggy images and trajectory heat-maps as input.

\begin{figure}
\centerline{
\includegraphics[width=0.975\columnwidth]{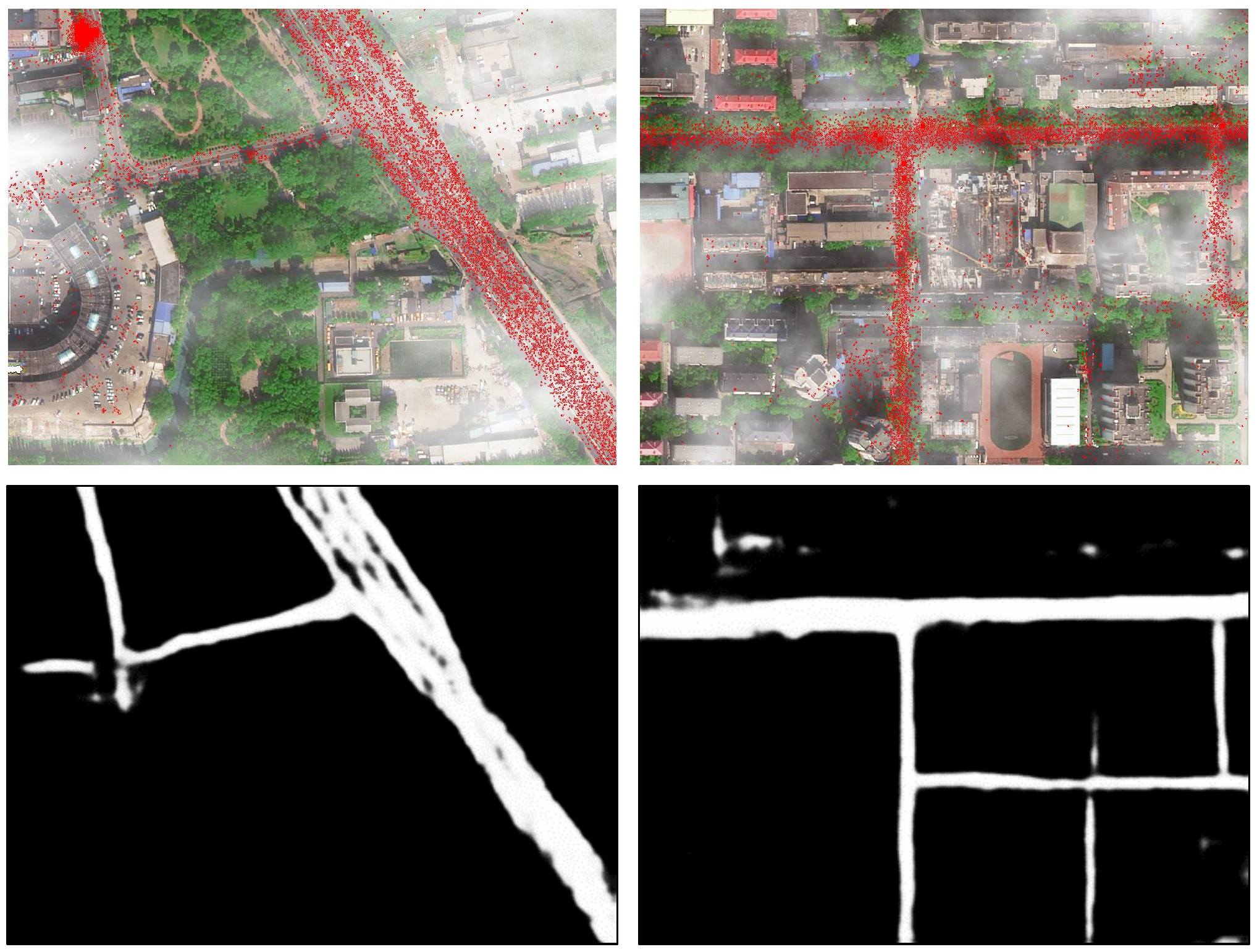}
}
\vspace{-3mm}
   \caption{The first row shows some foggy images and mass vehicle trajectories on the testing set of the foggy BJRoad dataset. Although traffic roads are occluded extremely in these images, our CMMPNet can still generate high-quality road network maps by fully exploiting the complementary information of vehicle crowdsourced trajectories, as shown in the second row.}
\vspace{0mm}
\label{fig:fog_result}
\end{figure}

\begin{table}[t]
  \caption{The performance of traffic road extraction based on foggy images and vehicle trajectories on the testing set of the foggy BJRoad dataset.}
  \vspace{-2mm}
\newcommand{\tabincell}[2]{\begin{tabular}{@{}#1@{}}#2\end{tabular}}
  \centering
    \begin{tabular}{c|c|c|c}
    \hline
    \multicolumn{1}{c|}{\textbf{Input}} & \textbf{Manner Way} & \textbf{A\_IoU (\%)} & \textbf{G\_IoU (\%)} \\
    \hline\hline
      Trajectory  & - & 52.38 & 52.90  \\
      Fog\_Img & - & 54.54 & 55.27  \\
      \hline
      \multirow{2}{*}{Fog\_Img + Trajectory}
       & Early Fusion & 57.98 & 58.49 \\
       & CMMPNet  & 60.45 & 61.06\\
    \hline
    \end{tabular}
  \label{tab:fog_image}
\end{table}

\begin{figure}[t]
\centerline{
\includegraphics[width=0.825\columnwidth]{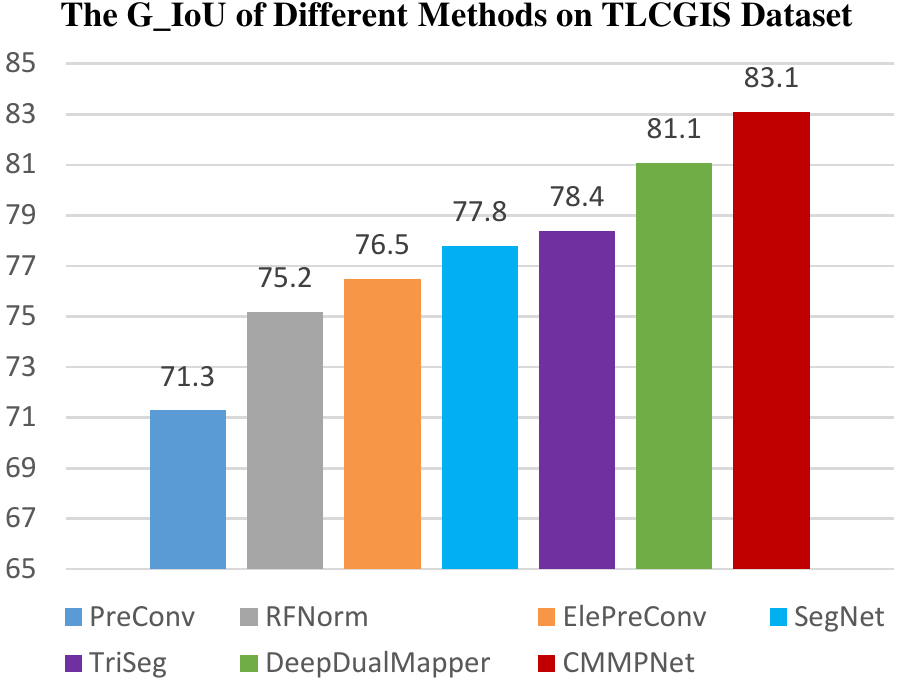}
}
\vspace{-3mm}
   \caption{The performance of different methods on the TLCGIS dataset. The proposed CMMPNet also outperforms all existing approaches for image+Lidar based road extraction.}
\vspace{1mm}
\label{fig:lidar_result}
\end{figure}

The results of all methods are summarized in Table~\ref{tab:fog_image}. We can observe that the unimodal D-LinkNet only obtains an A\_IoU 54.54\% and a G\_IoU 55.27\% when only using foggy images. Compared with the corresponding model using cloudless images, this model has a dramatic drop in performance, since traffic roads may be invisible in foggy images. When utilizing foggy images and trajectories simultaneously, the early fusion model obtains an A\_IoU 57.98\% and a G\_IoU 58.49\%. Based on the same D-LinkNet, our CMMPNet achieves a competitive A\_IoU 60.45\% and G\_IoU 61.06\%, having a performance improvement of at least 3\% compared with other models. Moreover, the visualizations in Fig. \ref{fig:fog_result} show that our CMMPNet can still generate high-quality road maps in foggy weather conditions. This is attributed to the fact that the vehicle trajectories can provide rich information to remedy the limitation of aerial images, and our method can fully capture their complementary information. In summary, crowdsourced trajectories are very crucial and beneficial for robust road extraction.

\begin{figure*}[t]
  \begin{center}
     \includegraphics[width=2\columnwidth]{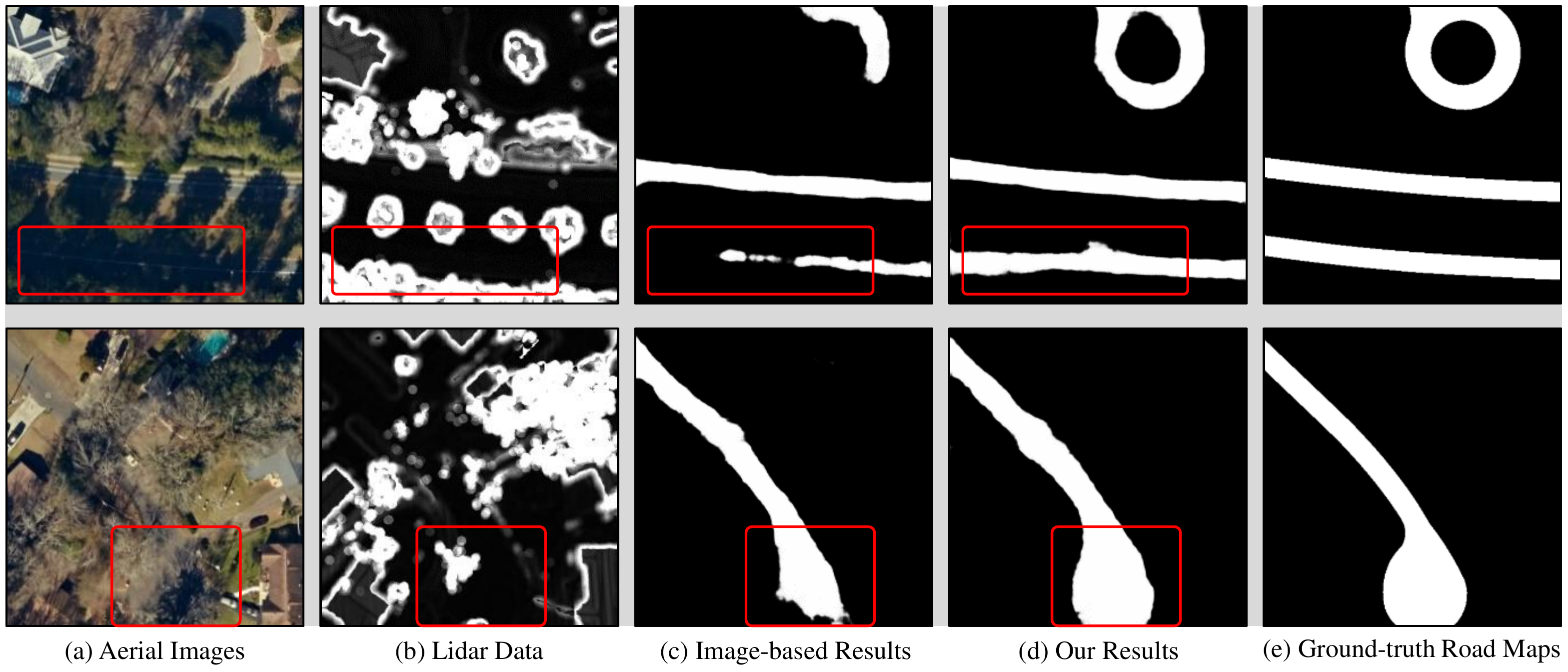}
  \vspace{-5mm}
  \end{center}
   \caption{Visualization of the generated traffic road maps on the TLCGIS dataset. (a) and (b) are the input aerial images and Lidar images. (c) are the results that only aerial images are taken as input. (d) are the results of our CMMPNet that utilizes both aerial images and Lidar data. (e) are the ground-truth road maps.}
\vspace{0mm}
\label{fig:lidar_map_visual}
\end{figure*}

\section{Apply to Image+Lidar based Extraction}\label{sec:lidar_experiment}
As mentioned above, our method is general for road extraction by exploiting multimodal information. In this section, we employ the proposed CMMPNet to recognize traffic roads from aerial images and Lidar data. As shown in Fig.~\ref{fig:lidar_map_visual}-(a,b), Lidar data can help to discover some occluded or inconspicuous roads in aerial images. Here we conduct extensive experiments on the TLCGIS~\cite{parajuli2018fusion} dataset, which consists of 5,860 pairs of aerial images and Lidar images rendered from raw Lidar point cloud data. The resolution of these images is 500$\times$500 and the geographical length of each pixel is 0.5 feet. This dataset is officially divided into training, test, and validation sets with each having 2,640, 2,400, and 240 samples respectively. On this dataset, we also take D-LinkNet as the backbone to develop our CMMPNet and optimize this model with the process described in Section \ref{sec:details}.

\subsection{Comparison with State-of-the-Art Metods}
In this subsection, we compare our CMMPNet with six state-of-the-art methods on the TLCGIS dataset. The details of these compared methods are described as follows.
{\bf{SegNet}} \cite{badrinarayanan2017segnet}: As a fully convolutional AutoEncoder, SegNet takes the concatenation of aerial images and Lidar images as input.
{\bf{PreConv}} \cite{parajuli2018fusion}: The Lidar images are first fed into a depth convolution unit (DepthCNN) implemented with two convolutional layers. The Lidar features and aerial images are then concatenated and fed into SegNet.
{\bf{RFNorm}} \cite{parajuli2018fusion}: Given aerial and Lidar images, some Random Forest classifiers~\cite{liaw2002classification} are first trained to estimate the road probability score at each location. The aerial-based and Lidar-based score maps are concatenated and fed into SegNet.
{\bf{ElePreConv}} \cite{parajuli2018fusion}: In this model, Lidar images are first encoded with two convolutions (8 and 4 filters), while aerial images are extended with an extra zero-initialized channel. The element-wise addition of 4-channel images and Lidar features are fed into FuseNet \cite{hazirbas2016fusenet}.
{\bf{TriSeg}} \cite{parajuli2018fusion}: This model consists of three SegNets. The first two SegNets respectively take aerial or Lidar images to generate the road probability maps, which are concatenated and fed into the third SegNet for final estimation.
{\bf{DeepDualMapper}} \cite{wu2020deepdualmapper}: This model has been described above and here we adopt D-LinkNet as the backbone network to reimplement this model.

The performance of all approaches is summarized in Fig. \ref{fig:lidar_result}. We can observe that the previous best-performing methods are TriSeg and DeepDualMapper, whose G\_IoU are 78.4\% and 81.1\%, respectively. Thanks to the cross-modal mutual refinement strategy, our CMMPNet achieves a new state-of-the-art G\_IoU 83.1\% on the TLCGIS dataset and greatly outperforms DeepDualMapper with an absolute improvement of 2\%. Moreover, we also visualize some results in Fig.~\ref{fig:lidar_map_visual}. As can be observed, the traffic road maps generated by our method are more accurate in complex scenarios. In summary, these quantitative and qualitative comparisons demonstrate that our CMMPNet is universal and effective to extract traffic roads from aerial images and Lidar data.

\begin{table}[t]
  \caption{The performance of different inputs and different representation learning manners on the testing set of TLCGIS dataset.}
  \vspace{-2mm}
\newcommand{\tabincell}[2]{\begin{tabular}{@{}#1@{}}#2\end{tabular}}
  \centering
    \begin{tabular}{c|c|c}
    \hline
    \multicolumn{1}{c|}{\textbf{Input}} & \textbf{Learning Manner} & \textbf{G\_IoU (\%)} \\
    \hline\hline
      Lidar  & - & 69.12 \\
      Image       & - & 80.96 \\
      \hline
      \multirow{3}{*}{Image+Lidar}
       & {~~}Late Fusion & 81.50 \\
       & Early Fusion    & 81.62 \\
       & CMMPNet  & 83.10 \\
    \hline
    \end{tabular}
  \label{tab:lidar_fusion}
\end{table}

\begin{table}[t]
  \caption{The influence of Non-Local Message Propagator and Gated Message Propagator on the testing set of TLCGIS dataset.}
  \vspace{-2mm}
\newcommand{\tabincell}[2]{\begin{tabular}{@{}#1@{}}#2\end{tabular}}
  \centering
    \begin{tabular}{ccc|c}
    \hline
    \textbf{Local} & \textbf{Global} & \textbf{Gate} & \textbf{G\_IoU (\%)} \\
    \hline\hline
    \checkmark & &            & 81.88 \\
    \checkmark & & \checkmark & 82.31 \\
    \hline
    \checkmark & \checkmark &            & 82.06 \\
    \checkmark & \checkmark & \checkmark & 83.10 \\
    \hline
    \end{tabular}
  \label{tab:lidar_global_gate}
  \vspace{2mm}
\end{table}

\subsection{Internal Analysis}
In this subsection, we verify the effectiveness of each component in the proposed CMMPNet for image+Lidar based road extraction. We first explore which manner can better exploit the information of these modalities. As shown in Table \ref{tab:lidar_fusion}, we can obtain a G\_IoU of 69.12\%, when only feeding the rendered Lidar images into D-LinkNet. When only using aerial images, the G\_IoU of D-LinkNet is 80.96\%, which indicates that aerial images are more important. Incorporating the information of aerial and Lidar images simultaneously, the early fusion model obtains a G\_IoU of 81.62\%, while the late fusion model has a comparable G\_IoU of 81.50\%. When fully exploring their complementary information with cross-modal message propagators, our CMMPNet achieves an impressive G\_IoU 83.10\%, outperforming the early/late fusion models with an absolute improvement of 1.5\%. This demonstrates that the proposed CMMPNet can also effectively capture the complementary information among aerial images and Lidar data.

We then explore the effect of global information and gate functions. Similar to Section \ref{sec:ablation}, we implement several variants of CMMPNet. As shown in Table \ref{tab:lidar_global_gate}, when only propagating local information with gate functions, ``Local+Gate'' model obtains a G\_IoU 82.31\%. When incorporating global information, ``Local+Global+Gate'' model has a better  G\_IoU 83.10\%, which indicates that the global information is also useful for image+Lidar based road extraction.
Moreover, by comparing the performance of ``Local+Global'' model and ``Local+Global+Gate'' model, we can observe that the gate functions help to make an absolute improvement of 1.04\% on G\_IoU, which also demonstrates the effectiveness of Gated Message Propagator for image+Lidar based road extraction.

\section{Conclusion}\label{sec:conclusion}
In this work, we investigate a challenging task for land remote sensing analysis, i.e., how to robustly extract traffic roads using the complementary information of aerial images and vehicle crowdsourced trajectories. To this end, we introduce a novel Cross-Modal Message Propagation Network (CMMPNet), which learns modality-specific features explicitly with two individual AutoEncoders and enhances these features mutually with a tailor-designed Dual Enhancement Module. Specifically, we comprehensively extract and dynamically propagate the complementary information of each modality to enhance the representation of another modality. Extensive experiments conducted on two real-world benchmarks show that the proposed CMMPNet is not only effective for image+trajectory based road extraction, but also suitable for image+Lidar based road extraction.

Nevertheless, there are still several issues worthy of further study. {\bf{First}}, the connectivity of traffic roads has not been explicitly explored in conventional works. Intuitively, the temporal information of vehicle trajectories could be utilized to distinguish disconnected road regions (e.g., urban roads are usually separated by fences and green belts). However, existing image+trajectory datasets lack the road connectivity annotation. To facilitate the researches in this field, we will construct a large-scale multimodal road extraction with rich connectivity annotation and propose a multimodal spatial-temporal framework to explicitly estimate the road connectivity in future work. {\bf{Second}}, some elevated roads at different heights are overlapped on aerial images. The height information accessed with GPS devices is relatively coarse. Thus in future work, we will also develop some advanced approaches to effectively recognize the roads at different heights with the coarse height information of crowdsourced trajectories.

\ifCLASSOPTIONcaptionsoff
  \newpage
\fi

\bibliographystyle{IEEEtran}
\bibliography{RoadSeg-Reference}

\begin{IEEEbiography}[{\includegraphics[width=1in,height=1.25in,clip,keepaspectratio]{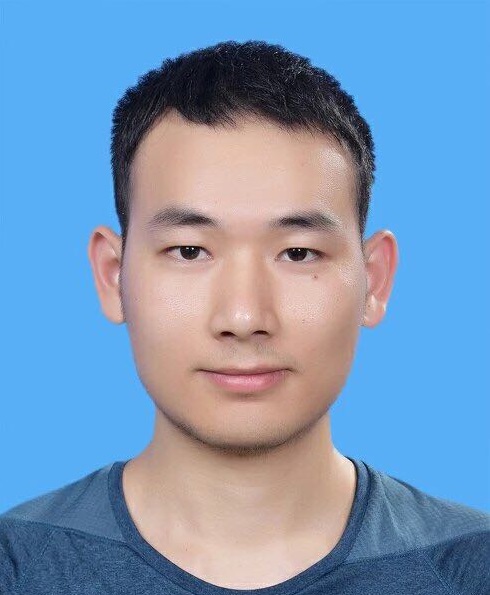}}]{Lingbo Liu}
received the Ph.D degree from the School of Computer Science and Engineering, Sun Yat-sen University, Guangzhou, China, in 2020. From March 2018 to May 2019, he was a research assistant at the University of Sydney, Australia. He is currently a postdoctoral fellow in
the Department of Computing at the Hong Kong Polytechnic University. His current research interests include machine learning and urban computing. He has authorized and co-authorized on more than 15 papers in top-tier academic journals and conferences. He has been serving as a reviewer for numerous academic journals and conferences such as TPAMI, TKDE, TNNLS, TITS, CVPR, ICCV and IJCAJ.
\end{IEEEbiography}

\begin{IEEEbiography}[{\includegraphics[width=1in,height=1.25in,clip,keepaspectratio]{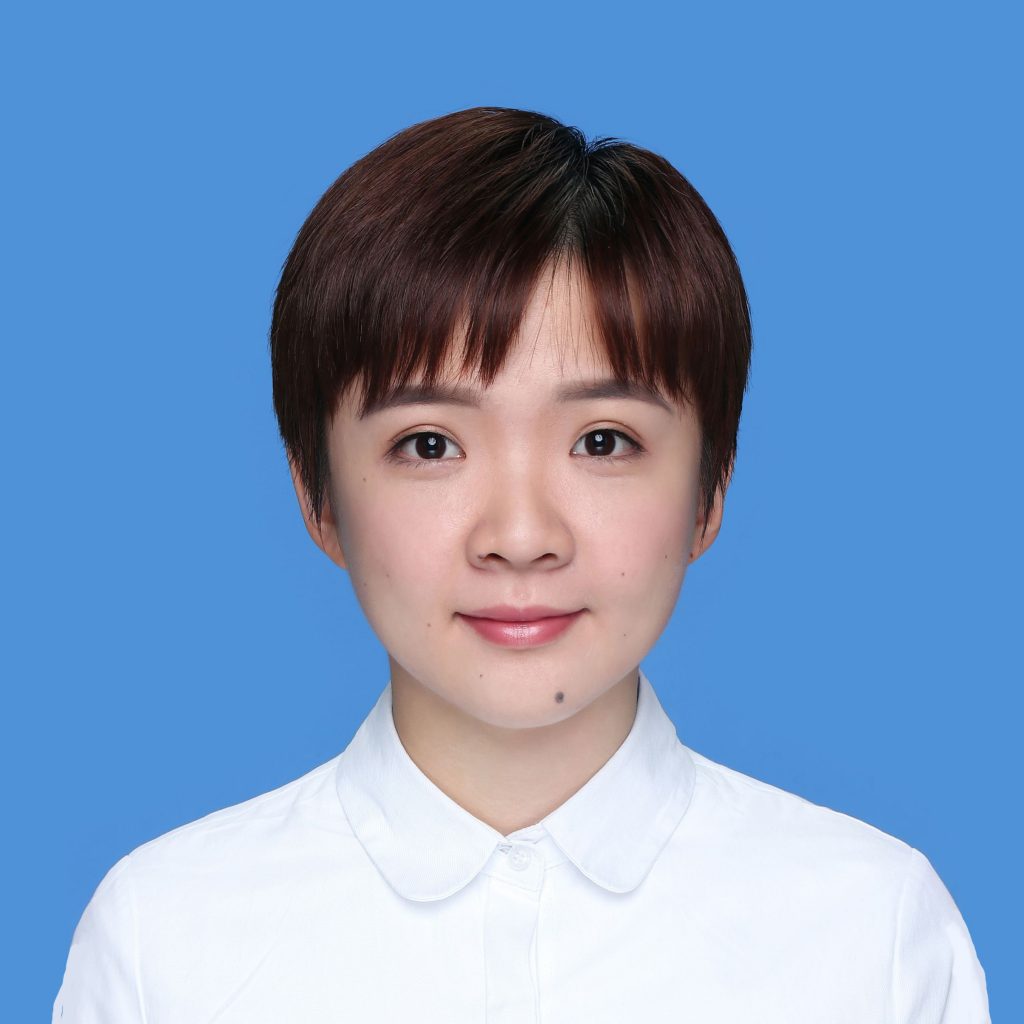}}]
{Zewei Yang}
received the B.E. degree from the School of Mathematics, Sun Yat-sen University, Guangzhou, China, in 2019, where she is currently pursuing the Master's degree in applied mathematics. His current research interests include machine learning and data mining.
\end{IEEEbiography}

\begin{IEEEbiography}[{\includegraphics[width=1in,height=1.25in,clip,keepaspectratio]{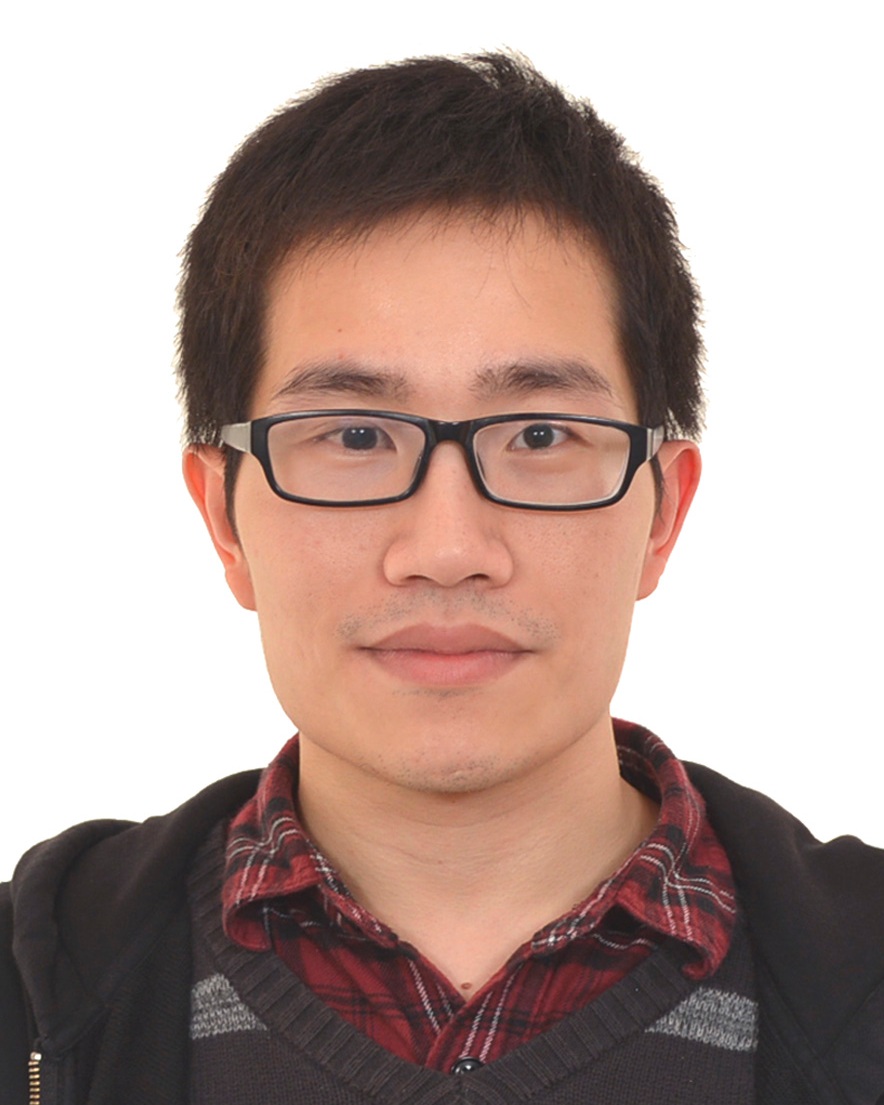}}]{Guanbin Li}
is currently an associate professor in School of Computer Science and Engineering, Sun Yat-sen University. He received his PhD degree from the University of Hong Kong in 2016. His current research interests include computer vision, image processing, and deep learning. He is a recipient of ICCV 2019 Best Paper Nomination Award. He has authorized and co-authorized on more than 60 papers in top-tier academic journals and conferences. He serves as an associate editor for journal of The Visual Computer, an area chair for the conference of VISAPP. He has been serving as a reviewer for numerous academic journals and conferences such as TPAMI, IJCV, TIP, TMM, TCyb, CVPR, ICCV, ECCV and NeurIPS.
\end{IEEEbiography}

\begin{IEEEbiography}[{\includegraphics[width=1in,height=1.25in,clip,keepaspectratio]{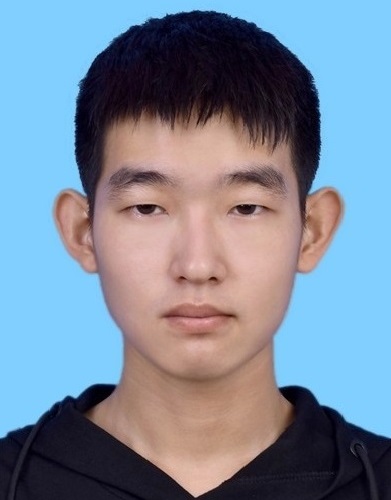}}]{Kuo Wang}
received the B.E. degree from the School of Computer Science and Engineering, Sun Yat-sen University, Guangzhou, China, in 2020, where he is currently pursuing the Ph.D degree in computer science. His current research interests include deep learning and recommended system.
\end{IEEEbiography}

\begin{IEEEbiography}[{\includegraphics[width=1in,height=1.25in,clip,keepaspectratio]{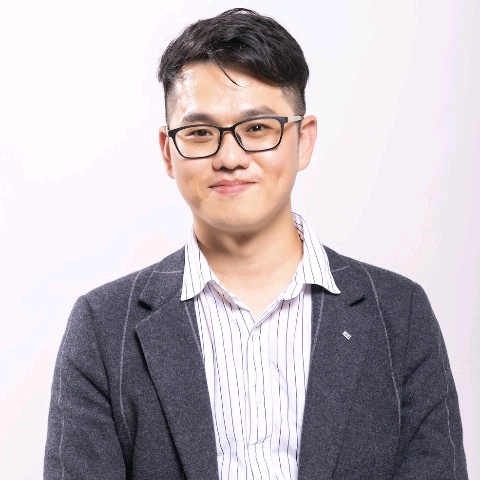}}]{Tianshui Chen}
received a Ph.D. degree in computer science at the School of Data and Computer Science Sun Yat-sen University, Guangzhou, China, in 2018. Before that, he received a B.E. degree from the School of Information and Science Technology. He is currently the lecturer in the Guangdong University of Technology. His current research interests include computer vision and machine learning. He has authored and coauthored approximately 20 papers published in top-tier academic journals and conferences. He has served as a reviewer for numerous academic journals and conferences, including TPAMI, IJCV, TIP, TMM, TNNLS, CVPR, ICCV, ECCV, AAAI, and IJCAI. He was the recipient of the Best Paper Diamond Award at IEEE ICME 2017.
\end{IEEEbiography}

\begin{IEEEbiography}[{\includegraphics[width=1in,height=1.25in,clip,keepaspectratio]{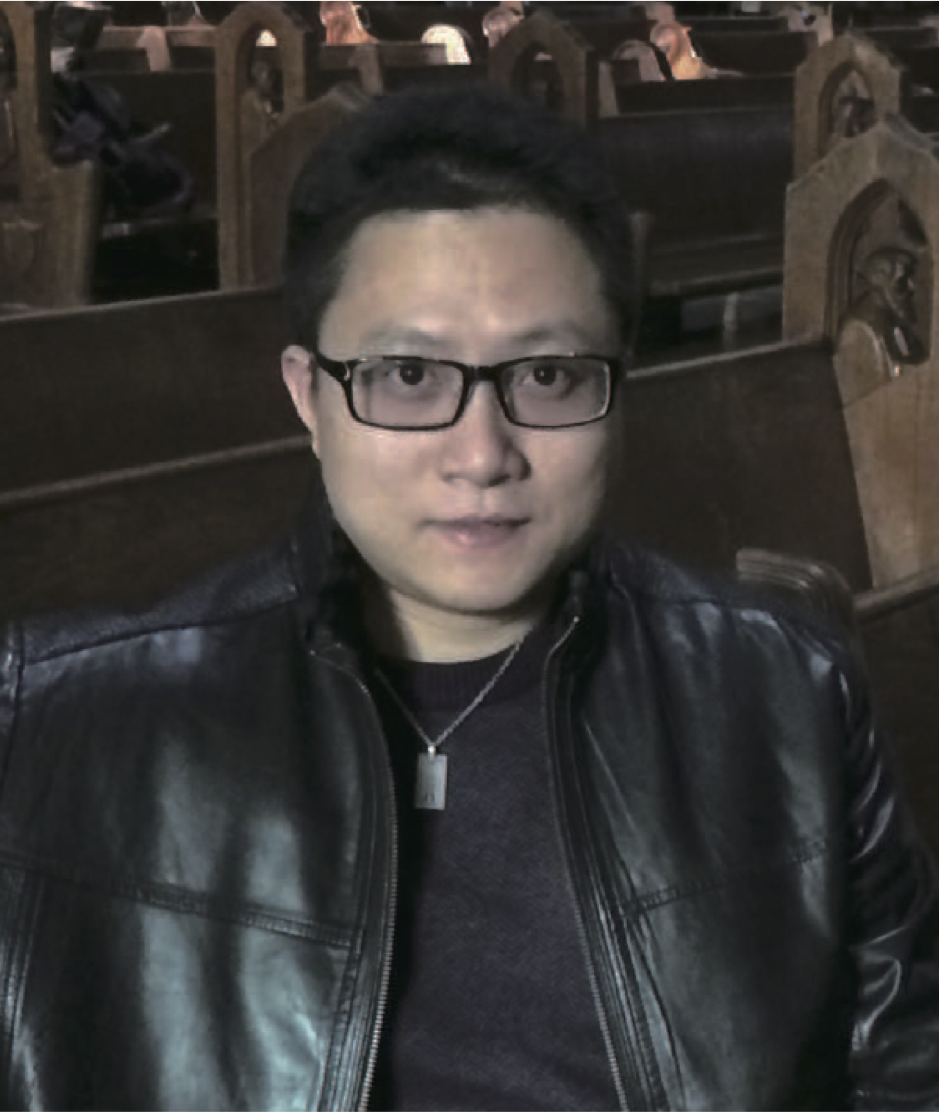}}]{Liang Lin}
is a full Professor of Sun Yat-sen University. He is the Excellent Young Scientist of the National Natural Science Foundation of China. From 2008 to 2010, he was a Post-Doctoral Fellow at University of California, Los Angeles. From 2014 to 2015, as a senior visiting scholar, he was with The Hong Kong Polytechnic University and The Chinese University of Hong Kong. From 2017 to 2018, he leaded the SenseTime R\&D teams to develop cutting-edges and deliverable solutions on computer vision, data analysis and mining, and intelligent robotic systems. He has authorized and co-authorized on more than 100 papers in top-tier academic journals and conferences. He has been serving as an associate editor of IEEE Trans. on Neural Networks and Learning Systems, IEEE Trans. Human-Machine Systems, The Visual Computer and Neurocomputing. He served as Area/Session Chairs for numerous conferences such as ICME, ACCV, ICMR. He was the recipient of Best Paper Nomination Award in ICCV 2019, Best Paper Runners-Up Award in ACM NPAR 2010, Google Faculty Award in 2012, Best Paper Diamond Award in IEEE ICME 2017, and Hong Kong Scholars Award in 2014. He is a Fellow of IET.
\end{IEEEbiography}

\end{document}